\documentclass[journal]{IEEEtran}
\usepackage{amsmath,amsfonts}
\usepackage{algorithmic}
\usepackage{array}
\usepackage[caption=false,font=footnotesize,labelfont=rm,textfont=rm]{subfig}
\usepackage{textcomp}
\usepackage{stfloats}
\usepackage{url}
\usepackage{verbatim}
\usepackage{graphicx}
\usepackage{bm}
\usepackage{multirow}
\usepackage[switch]{lineno}
\usepackage{balance}
\usepackage{algorithm}
\usepackage{cite}
\usepackage{booktabs}
\usepackage{subfig} 
\usepackage{array}
\usepackage{flushend}

\def\BibTeX{{\rm B\kern-.05em{\sc i\kern-.025em b}\kern-.08em
		T\kern-.1667em\lower.7ex\hbox{E}\kern-.125emX}}
	
\begin{document}
\title{Variational Disentangled Graph Auto-Encoders for Link Prediction}
\author{Jun Fu$^*$, \IEEEmembership{Senior Member, IEEE,} Xiaojuan Zhang, Shuang Li, Dali Chen
	\thanks{This work was supported in part by the National Natural Science Foundation of China under Grants 61825301.}
		\thanks{Jun Fu is the corresponding author and with the State Key Laboratory of Synthetical Automation for Process Industries, Northeastern University, Shenyang, China. (e-mails: junfu@mail.neu.edu.cn)}
		\thanks{Xiaojuan Zhang is with the State Key Laboratory of Synthetical Automation for Process Industries, Northeastern University, Shenyang, China. (e-mails: xiaojuanzhang123@gmail.com).}
		\thanks{Shuang Li is with School of Computer Science and Technology, Beijing Institute of Technology, Beijing, China. (e-mail: shuangli@bit.edu.cn)}
		\thanks{Dali Chen is with the College of Information Science and Engineering, Northeastern University, Shenyang, China (e-mail: chendali@ise.neu.edu.cn).}
}

\markboth{Journal of \LaTeX\ Class Files,~Vol.~18, No.~9, September~2020}%
{How to Use the IEEEtran \LaTeX \ Templates}

\maketitle

\begin{abstract}
With the explosion of graph-structured data, link prediction has emerged as an increasingly important task. Embedding methods for link prediction utilize neural networks to generate node embeddings, which are subsequently employed to predict links between nodes. However, the existing embedding methods typically take a holistic strategy to learn node embeddings and ignore the entanglement of latent factors. As a result, entangled embeddings fail to effectively capture the underlying information and are vulnerable to irrelevant information, leading to unconvincing and uninterpretable link prediction results. To address these challenges, this paper proposes a novel framework with two variants, the disentangled graph auto-encoder (DGAE) and the variational disentangled graph auto-encoder (VDGAE). Our work provides a pioneering effort to apply the disentanglement strategy to link prediction. The proposed framework infers the latent factors that cause edges in the graph and disentangles the representation into multiple channels corresponding to unique latent factors, which contributes to improving the performance of link prediction. To further encourage the embeddings to capture mutually exclusive latent factors, we introduce mutual information regularization to enhance the independence among different channels. Extensive experiments on various real-world benchmarks demonstrate that our proposed methods achieve state-of-the-art results compared to a variety of strong baselines on link prediction tasks. 
Qualitative analysis on the synthetic dataset also illustrates that the proposed methods can capture distinct latent factors that cause links, providing empirical evidence that our models are able to explain the results of link prediction to some extent. All code will be made publicly available upon publication of the paper.
\end{abstract}

\begin{IEEEkeywords}
Link prediction, disentangled representation learning, mutual information, variational graph auto-encoder
\end{IEEEkeywords}

\section{Introduction}
\IEEEPARstart{R}{ecent} years have witnessed an impressive rise and expansion of graph-structured data, consisting of nodes and edges (links) that connect nodes. Graph-structured data is widely used to model relationships, interactions, and dependencies among entities in various domains such as social networks, biological networks, and traffic networks. Although numerous links within the real-world graphs have been observed, a comprehensive observation remains incomplete. Consequently, link prediction, which aims to predict the existence of edges between pairs of nodes, has emerged as one of the research hotspots in graph domain.

Link prediction finds extensive practical application in many real-world graphs. In social networks, predicting potential friendships or collaborations can facilitate community detection \cite{yang2013community} and targeted marketing \cite{Zhang2010}. In biological networks, link prediction aids in understanding protein-protein interactions \cite{Lv2021} and gene regulatory networks \cite{Karlebach2008}. Moreover, in recommender systems, predicting user-item links enables personalized recommendations \cite{Xia2022} and enhances user experience \cite{10.1145/3397271.3401137}. In information retrieval, link prediction contributes to crucial tasks such as web page ranking \cite{Brin1998} and knowledge graph completion \cite{Ji2022}. The demands in these domains drive the development of link prediction methods and techniques to provide more accurate predictions and yield improved outcomes for various applications.

\begin{figure}[t!]
	\centering
	\includegraphics[width=1\linewidth]{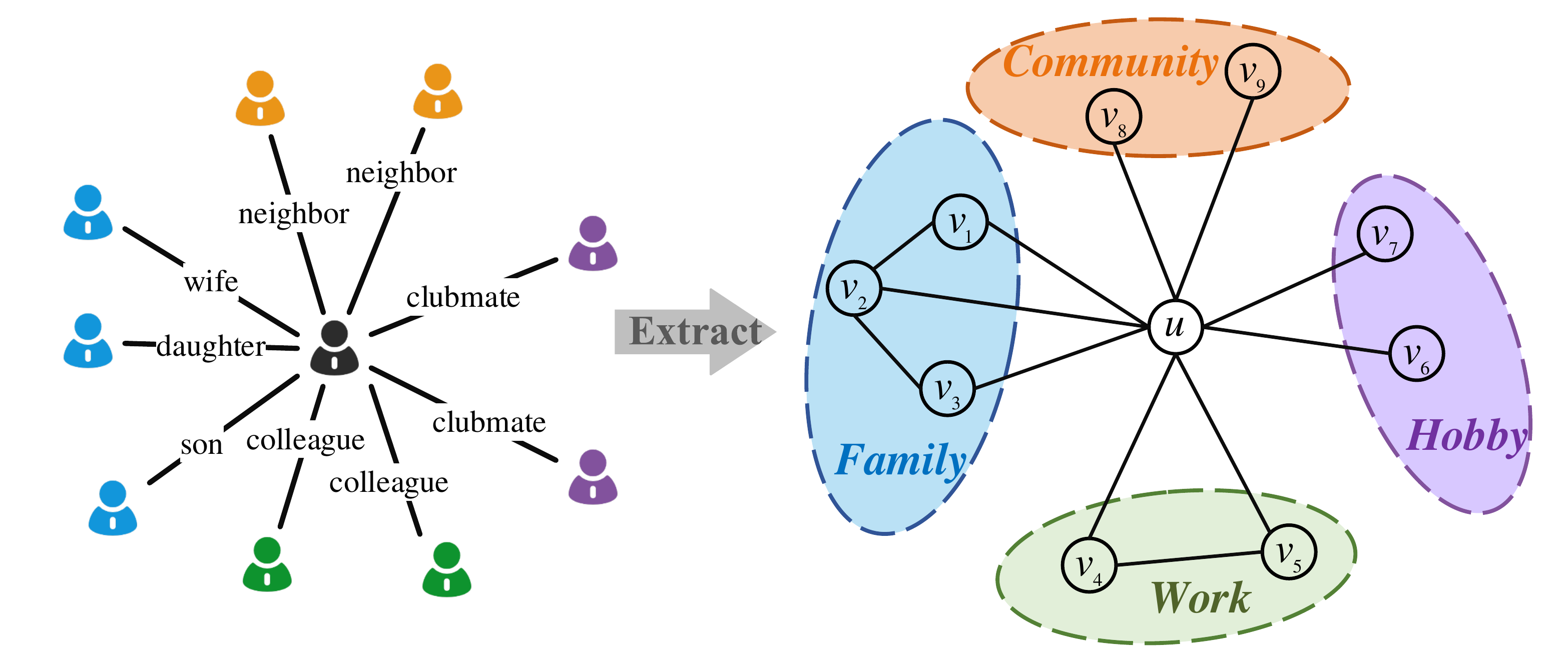}
	\caption{An example on social network, where node $u$ represents a specific person, and other nodes $v_{i},i=1,\cdots,9$ represents neighbors of node $u$. It can be seen that the social network is inherently heterogeneous: an individual may connect with others due to four latent factors: work, hobbies, family, and community.}
	\vspace{-1em}
	\label{fig:subgraph}
\end{figure}
Existing methods for link prediction broadly fall into two categories: heuristic methods and embedding methods.
Most heuristic methods estimate the likelihood of links based on similarity scores between nodes \cite{Lue2011,Barabasi1999,Brin2012,Jeh2002,Zhang2017,Adamic2003,Jaccard1902,Newman2001,Zhou2009,Katz1953}. However, these methods often rely on strong assumptions and may not leverage node attribute information, limiting their ability to capture complex patterns in large and complex real-world graphs. Embedding methods are based on the encoder-decoder framework \cite{Kipf2016,kingma2013auto,DBLP:conf/uai/DavidsonFCKT18}. The encoder distills high-dimensional information of both graph structure and graph attribute into node embeddings. The decoder takes the node embeddings to predict links. Early work utilized shallow embedding approaches, such as factorization-based methods \cite{10.1145/2488388.2488393,10.1145/2806416.2806512,10.1145/2939672.2939751} and random walk methods \cite{10.1145/2623330.2623732,Tang2015,10.1145/2939672.2939754,Ribeiro2017}.  However, the lack of parameter sharing between nodes in the encoder renders shallow embedding methods statistically and computationally inefficient. More recently, graph neural networks (GNNs) \cite{kipf2017semisupervised,Hamilton2017,DBLP:conf/iclr/VelickovicCCRLB18}, based on spectral graph theory and graph Fourier transform \cite{spectral}, have emerged as deep embedding methods for link prediction.
GNNs have shown promising link prediction results due to their high scalability and adaptability to graphs of different sizes. 

Despite the huge success, most existing embedding methods for link prediction take a holistic strategy to learn entangled node embeddings: the embedding for a node is learned by indiscriminately aggregating all its local neighbors. However, the entangled node embedding poses two challenges for link prediction: (1) The entangled node embedding usually yields link prediction outcomes that lack interpretability and robustness. Real-world graphs are inherently heterogeneous, as link formation is influenced by diverse latent factors. However, entangled node embeddings lack the ability to identify the underlying latent factors that cause links in the graph.
As shown in Fig. \ref{fig:subgraph}, there are four latent factors that enable an individual to establish links with others: work, hobbies, family, and community. When predicting links that are specifically related to family, node embeddings should focus more on the three nodes ($v_{1}$, $v_{2}$, $v_{3}$) associated with family. However, entangled embeddings treat all four latent factors indiscriminately, leading to limited interpretability and poor robustness of link prediction results. (2) The entangled node embedding gives rise to a decrease in the accuracy of link prediction due to excessive interaction with irrelevant information. For instance, when nodes are embedded with noise or interference components from the node attribute information, these error signals may lead to incorrect link prediction results. In summary, the holistic strategy fails to distinguish and capture different latent factors and thus generates entangled embeddings. Such embeddings are highly likely to produce unconvincing and uninterpretable link prediction results.
\begin{figure*}[t!]
	\centering
	\includegraphics[width=1\linewidth]{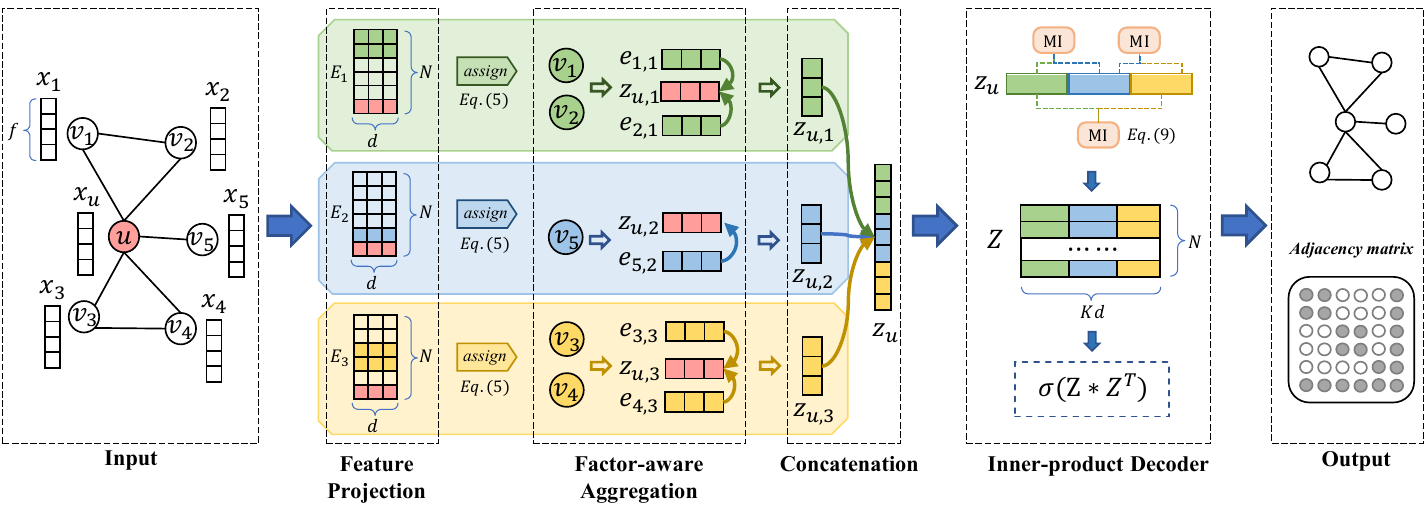}
	\caption{The architecture overview of our framework. The proposed framework takes node $u$, its neighbors $v_{i},i=1,\cdots,5$, and their feature vectors $x_{i} \in \mathbb{R}^{f}$ as \textbf{input}. 
	The disentangled graph encoder contains three key modules: (1) \textbf{feature projection} projects the feature vector $x_{i}$ into $K$ channels to generate the initial node embedding matrix $E_{k}=[e_{1,k}, \cdots, e_{N,k}]$ for the $k$-th channel; (2) \textbf{factor-aware aggregation} assigns neighbors $v_{i}$ to different channels by Eq.\eqref{probability} and perform channel-specific aggregation to update the embedding vector $z_{u,k}$ of the $k$-th channel; (3) the disentangled node embedding $z_{u} \in \mathbb{R}^{Kd}$ is the \textbf{concatenation} of total $K$ channels. After that, MI regularization is introduced to enhance the independence between the disentangled components. The above operations are performed on all $N$ nodes in the graph to obtain the node embedding matrix $Z \in \mathbb{R}^{N \times Kd}$. Finally, the \textbf{inner product decoder} conducts the reconstruction of the \textbf{adjacency matrix} to perform link prediction. The framework is optimized by Eq.\eqref{L0} for GDAE or Eq.\eqref{L1} for VGDAE.}
	
	\label{fig:framework}
	\vspace{-1em}
\end{figure*}

To address these challenges, we are the pioneers to provide insights on how the disentanglement strategy can be leveraged to enhance the performance of link prediction tasks, which remains largely unexplored in this field. In this work, we propose a novel framework comprising two different variants, disentangled graph auto-encoder (DGAE) and variational disentangled graph auto-encoder (VDGAE), to perform link prediction task on graph-structured data (see Fig. \ref{fig:framework}). 
Our proposed framework consists of a disentangled graph encoder that generates disentangled node embedding, and an inner product decoder that performs link prediction tasks. Specifically, the encoder projects the node feature vector into various low-dimensional channels to generate initial node embeddings. Furthermore, the encoder automatically identifies the neighbors of nodes and assigns these neighbors to the corresponding channels to conduct factor-aware aggregation, thereby avoiding interference from irrelevant information. The embedding vectors from multiple different channels are concatenated to obtain the final disentangled node embedding. Through the above operations, each channel can capture semantic information corresponding to various latent factors that cause links. Finally, the inner product decoder utilizes the disentangled node embeddings to reconstruct the adjacency matrix and perform link prediction.

The main challenge that affects the quality of disentangled node embeddings is how to avoid different channels capturing the same latent factor, that is, to ensure independence among different channels. The key insight is that each channel should capture a unique latent factor. To address the challenge, we employ mutual information (MI) to measure the independence among channels and impose regularization constraints by minimizing MI. Specifically, we calculate the MI between every two channels and incorporate it into the loss function, encouraging the channels to prioritize learning mutually exclusive semantic information during optimization.  In this way, the disentangled node embeddings are more likely to capture comprehensive latent factors from the original attribute data and reflect the heterogeneity of real-world graphs. Moreover, various latent factors influencing link formation also contribute to explaining the results of link prediction to some extent.
 
Furthermore, we conduct extensive experiments on 18 real-world graphs to demonstrate the performance of the proposed methods for link prediction. Experimental results validate the superiority of our proposed methods, where DGAE and VDGAE achieve state-of-the-art performance in link prediction compared to a variety of competitive baselines. Moreover, we perform abundant quantitative and qualitative experiment analyses. The ablation study and sensitivity analysis confirm the effectiveness of our proposed models. 
Finally, we generate a synthetic dataset to conduct disentanglement analysis and visualization of node embeddings. The results reveal that the proposed models indeed learn disentangled node embeddings which are mutually independent across different channels.

Our main contributions are summarized as follows:
\begin{itemize}
	\item {\textbf{Link prediction in the disentanglement manner.} 
		\\ We are the pioneers to apply the disentanglement strategy to link prediction, and propose two novel methods, namely GDAE and VDGAE. Our methods excel at identifying the latent factors of real-world graphs, allowing them to learn highly expressive disentangled node embeddings, which result in significant performance enhancements for link prediction tasks.}
	\item {\textbf{Independence constraints among channels.} 
		\\ We propose MI regularization to enforce the independence among different channels so that various channels can capture mutually exclusive information. }
	\item {\textbf{Evaluation on real-world and synthetic datasets.} 
		\\ We conduct extensive experiments on 18 challenging real-world datasets to verify the effectiveness of DGAE and VDGAE. Our proposed methods achieve state-of-the-art results against a variety of baselines. Moreover, we generate a synthetic dataset to provide empirical evidence that our models successfully disentangle the node representations and capture the latent factors that cause links.}
\end{itemize}
\section{Related Work}
Our methods draw inspiration from the field of representation learning on graphs, conceptually, related to variational graph auto-encoder and graph disentangled representation learning. In what follows, we provide a brief overview on related work in both fields.
\subsection{Variational Graph Auto-Encoder}
Inspired by the advancement of deep learning techniques, a large number of VGAE-based approaches for link prediction have emerged in recent years. Based on variational auto-encoder (VAE) \cite{kingma2013auto}, the prominent attempt is variational graph auto-encoder (VGAE) \cite{Kipf2016}, a general framework that leverages a graph convolutional network (GCN) encoder and an inner product decoder. VGAE tremendously improves the performance of link prediction on three network citation datasets.
Later, this idea is adopted and improved by subsequent works.  Hyperspherical VAE (s-VAE)  \cite{DBLP:conf/uai/DavidsonFCKT18} replaces the assumption that the prior distribution of VGAE is the normal distribution with von Mises-Fisher (vMF) in hyperspherical space.
Adversarially regularized variational graph auto-encoder (ARVGA)  \cite{DBLP:conf/ijcai/PanHLJYZ18} enforces the node embeddings to match the prior distribution by an adversarial training scheme.
Unlike the previous methods, Graphite-VAE \cite{Grover2019} adopts an iterative GNN-based strategy to augment the decoder.
Recently, variational graph normalized auto-encoder (VGNAE) \cite{10.1145/3459637.3482215} derives better embeddings for isolated nodes by way of adding an $L2$-normalization layer to the encoder. However, in the above methods, the node embeddings lack the exploration of latent factors. Compared with these works, our proposed methods take into full account the inherent heterogeneity of real-world graphs and employ a disentanglement strategy to learn node embeddings. 

\subsection{Graph Disentangled Representation Learning}
Recently, the huge success of disentangled representation learning in the computer vision domain \cite{GonzalezGarcia2018,9241434} has attracted a lot of interest in graph-structured data. 
Graph disentangled representation learning aims to learn low-dimensional embeddings by decomposing the representations of entities (nodes, edges, subgraphs) on a graph into multiple explainable components so that the embeddings can be generalized to unseen nodes and benefit downstream tasks.
However, works on graph disentangled representation learning are rather limited. Pioneering attempts occur on node-level disentanglement. Disentangled graph convolutional network (DisenGCN) \cite{Ma2019} employs the neighborhood routing mechanism that can automatically identify latent factors to learn disentangled node representations. Furthermore, a wide variety of independence constraints are imposed on the initial graph disentangled framework between different disentangled components.
Independence promoted graph disentangled networks (IPGDN) \cite{Liu2020} incorporates Hilbert-Schmidt Independence Criterion (HSIC) into DisenGCN to promote independence between the latent representations. Adversarial disentangled graph convolutional network (ADGCN) \cite{Zheng2021} utilizes an adversarial regularizer to improve the separability among different latent factors. These independence constraints provide considerable gains over the semi-supervised node classification tasks. 
Then several networks are proposed to conduct graph-level disentanglement and edge-level disentanglement. Factorizable graph convolutional network (FactorGCN) \cite{Yang2020} disentangles the whole complete graph into several multi-relation subgraphs to yield disentangled embeddings. 
It seems that the existing methods, whether at the node-level, edge-level, or graph-level, all concentrate on the semi-supervised node classification tasks and achieve fairly satisfactory results. In contrast, our work provides insights on utilizing the disentanglement strategy to enhance the performance of link prediction tasks, an area that has received relatively limited exploration.

Rather than training models based upon labels, another line of work develops disentangled embeddings by employing self-supervised learning strategies, including the design of heuristics pretext tasks \cite{Zhao2022} and the incorporation of contrastive learning \cite{Li2021}. Research also has surged on disentangled representation learning over heterogeneous graphs, which contain multi-typed nodes and multi-relation edges. The majority of these methods fall into two broad categories: methods that are designed for knowledge graphs \cite{10.1145/3459637.3482424,Yu2022,10.1145/3534678.3539453} and for recommender systems \cite{10.1145/3397271.3401137,10.1145/3340531.3411996}, which involve not only node-level information but also edge-level information of the graph. Our work is conceptually inspired by these methods. 
However, unlike these previous approaches, our work provides a new perspective on utilizing mutual information to constrain the independence among channels and capturing multiple underlying latent factors. Consequently, the proposed methods facilitate the generation of disentangled node embeddings.
\section{Preliminary}

\subsection{Notations}
\begin{figure}[t!]
	\centering
	\includegraphics[width=1\linewidth]{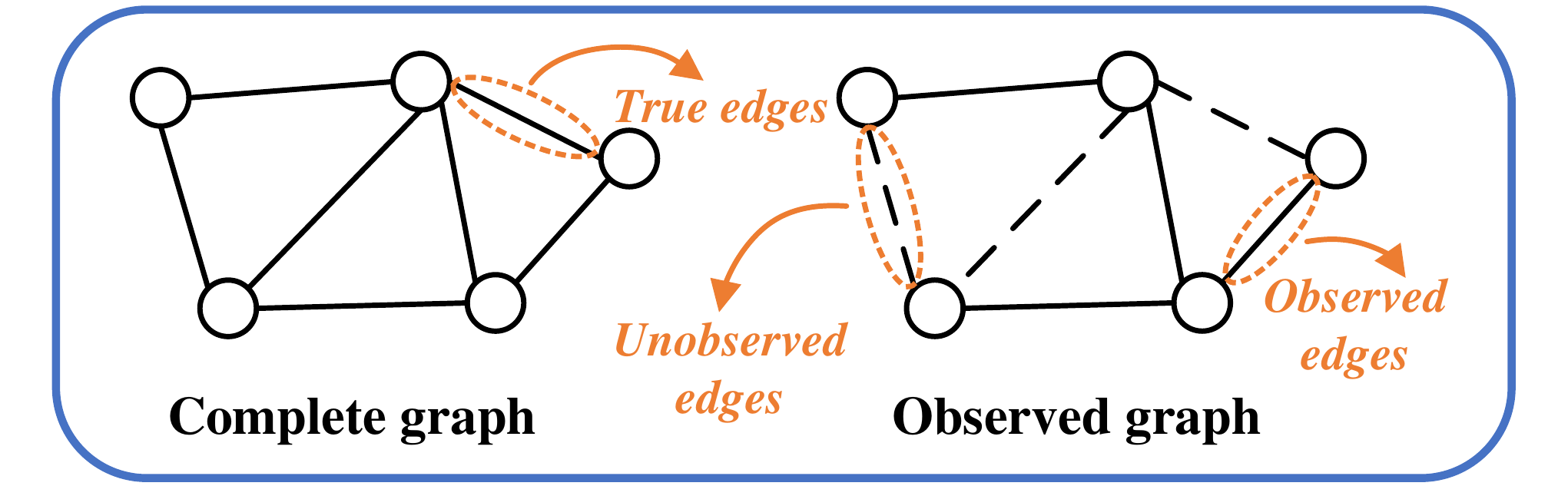}
	\caption{An illustration of the observed graph and the complete graph.}
	\label{notation}
	\vspace{-1em}
\end{figure}

Consider an undirected, unweighted graph $G=(V,E)$, where $V$ denotes the vertex set comprised of $N$ nodes and $E$ is the set of edges. Since complete graphs in the real world are not fully observed, we cannot obtain all true edges. We make the following definition.
\newtheorem{definition}{Definition}
\begin{definition}[The observed graph]
	The observed graph is defined as $G_{o}=(V,E_{o})$, where $V$ denotes the vertex set comprised of $N$ nodes and $E_{o}$ is the set of observed edges.
\end{definition}
\begin{definition}[The complete graph]
	The complete graph is defined as $G_{c}=(V,E_{c})$, where $V$ denotes the vertex set comprised of $N$ nodes and $E_{c}$ is the set of all true edges.
\end{definition}
The observed edge set $E_{o}$ is a subset of true edge set $E_{t}$, i.e. $E_{o} \subset  E_{t}$. We also define the a candidate set $E_{ca}$ with $|E_{c}| \leq |E_{ca}| \leq |N*(N-1)/2|$, which consists of both true edges and false edges. 
Moreover, each node has a feature vector $x_{i} \in \mathbb{R}^{f}$, and $X=\left [x_{1},\cdots,x_{N}\right ]\in \mathbb{R}^{N\times f}$ represents the feature matrix that encodes all node attribute information. The graph $G_{o}$ can be represented as an adjacency matrix $A\in \mathbb{R}^{N\times N}$ with $A_{i,j}=1$ if $(i,j) \in E_{o}$, otherwise $A_{i,j}=0$. An illustration of the observed graph and the complete graph is shown in Fig. \ref{notation}.
\subsection{Problem Formulation}
\newtheorem{problem}{Problem}
\begin{problem}[Link Prediction] 
	The goal of the link prediction task is to train an edge classifier $g_{\psi}\left (\cdot \right )$ with the learnable parameters $ \psi $ that can determine whether the edge in the candidate set $e_{i,j} \in E_{ca}$ belongs to the complete edge set $E_{c}$. Inputing $ e_{i,j} \in E_{ca}$, the edge classifier is defined as follows:
	\begin{equation}
		g_{\psi}\left (e_{i,j} \right )=
		\begin{cases}
			0 & \text{ if } e_{i,j} \notin E_{c} \\ 
			1 & \text{ if } e_{i,j} \in  E_{c}
		\end{cases}
		\label{classifier}
	\end{equation}	
\end{problem}
Our entire framework $g_{\psi}\left (\cdot \right )$ for link prediction consists of a disentangled graph encoder $f_{\theta }\left ( \cdot  \right )$ with the learnable parameters $\theta$, and an inner product decoder $y\left (\cdot \right )$ with no parameters. The disentangled graph encoder is devoted to generating low-dimensional disentangled embeddings, which can be restored to an adjacency matrix $A\in \mathbb{R}^{N\times N}$ corresponding to the graph structure by the decoder. The candidate set here is $E_{ca}: V \times  V$, which contains a total of $|N*(N-1)/2|$ edges (total node-pairs). The input of the encoder is the adjacency matrix $A$ and the feature matrix $X$. Hence, the proposed framework can be further formalized as:
\begin{align}
	\theta^{*}=&\mathop{\arg\min}\limits_{\psi}\mathcal{L}_{BCE}(g_{\psi}\left (e_{i,j} \right )) \notag \\
	=&\mathop{\arg\min}\limits_{\theta}\mathcal{L}_{con}(f_{\theta }(A, X),y),
	\label{framework}
\end{align}
where $\mathcal{L}_{BCE}$ is the binary cross-entropy loss for the edge classifier. Specifically, for our VGAE-based framework, we utilize the reconstruction loss function $\mathcal{L}_{con}$ (for the specific form, see \ref{Optimization}).

Moreover, for the disentangled graph encoder, we assume that the learned node embeddings consist of $K$ components corresponding to $K$ latent factors. In detail, for a given node $u$ with the feature vector $x_{u} \in \mathbb{R}^{f}$, the embedding vector is $z_{u}=\left[ z_{u,1}, z_{u,2}, \cdots, z_{u,K} \right]$, where $z_{u} \in \mathbb{R}^{Kd}$, $d \leq f$. $z_{u,k}\in \mathbb{R}^{d}$ denotes the embedding related to $k$-th latent factors, namely $k$-th channel.
\section{VDGAE: the Proposed Model}
In this section, we present the overall framework with two variants, termed disentangled graph auto-encoder (DGAE) and variational disentangled graph auto-encoder (VDGAE), an end-to-end framework that learns disentangled node embeddings for link prediction. The architecture overview of the proposed framework is shown in Fig. \ref{fig:framework}. 
\subsection{Disentangled Graph Encoder}
The disentangled graph encoder aims to learn the disentangled node embeddings, including three key modules: feature projection, factor-aware aggregation, and concatenation. The overall disentangled graph encoder is schematically depicted in Algorithm \ref{alg:algorithm}.
\subsubsection{Feature Projection}
In order to learn disentangled node embeddings, we assume that the node feature space can be decomposed into $K$ independent components associated with $K$ different latent factors. The value of $K$ is expected to vary across different graphs, as the number of latent factors can differ for different graphs. 
Moreover, it is crucial to carefully choose the value of parameter $K$, as a smaller value may not effectively capture all the latent factors, while a larger value leads to increased computational complexity. 
Here we linearly project the feature vector $x_{u}$ into different subspaces to initialize the embedding vector for each channel:
\begin{equation}
	e_{u,k}= W_{k}^{\top}x_{u}+b_{k} \label{feat_projection} ,
\end{equation}	
where $W_{k}\in \mathbb{R}^{f\times d}$ and $b_{k}\in \mathbb{R}^{d}$ are the learnable parameters related to $k$-th channel. Moreover, to ensure numerical stability, we impose $L2$ regularization on $e_{u,k}$: 
\begin{equation}
	e_{u,k}=\frac{e_{u,k}}{\left \| e_{u,k} \right \|_{2}} \label{embed_norm}.
\end{equation}	
By performing the same operation (Eq.\eqref{feat_projection}, Eq.\eqref{embed_norm}) on all node feature vectors, the initial node embedding matrix for $k$-th channel is obtained: $E_{k}=[e_{1,k}, \cdots, e_{N,k}]$, $E_{k} \in \mathbb{R}^{N \times d}$.
\begin{algorithm}[tb]
	\caption{Disentangled Graph Encoder}
	\label{alg:algorithm}
	\textbf{Input}: the feature vector $x_{u}\in \mathbb{R}^{f}$, its neighbors’ features $\left\{x_{v}\in \mathbb{R}^{f}:(u,v) \in G \right\}$, channel numbers $K$,  routing iterations $T$, $\phi$-VAN iterations $M$\\
	\textbf{Parameter}: $W_{k}\in \mathbb{R}^{f\times d}$, $b_{k}\in \mathbb{R}^{d}$, $k=1,\cdots ,K$ \\
	\textbf{Output}: node disentangled representations $z_{u} \in \mathbb{R}^{Kd}$ \par   
	\begin{algorithmic}[1] 
		\FOR{$i \in \left \{ u \right \}\cup \left \{ v:(u,v)\in G \right \}$}
		\FOR{$k=1,\cdots ,K$}
		\STATE Calculate $e_{u,k}$ by Eq.\eqref{feat_projection}
		\STATE $e_{u,k} \leftarrow e_{u,k}/\left \| e_{u,k} \right \|_{2}$
		\ENDFOR
		\ENDFOR	
		\STATE $z_{u,k} \leftarrow e_{u,k}$, $k=1,\cdots ,K$
		\FOR{routing iterations $t=1,\cdots,T$}
		\FOR{$v$ that satisfies $(u,v)\in G $}
		\STATE Calculate $p_{v,k}^{(t)}$ by Eq.\eqref{probability}
		\ENDFOR
		\FOR{channel $k=1,\cdots ,K$}
		\STATE Update $z_{u,k}^{(t)}$ by Eq.\eqref{aggregation}
		\STATE $z_{u,k}^{(t)} \leftarrow z_{u,k}^{(t)}/ \left \| z_{u,k}^{(t)} \right \|_{2}$
		\ENDFOR
		\ENDFOR	
		\FOR{channels $i,j \in K$ ($i\neq j$)}
		\STATE Calculate MI between every two different \\
		channels of node representations by Eq.\eqref{Lmi}
		\FOR{$\phi$-VAN iterations $m=1,\cdots,M$}
		\STATE Update $\phi $ by minimize Eq.\eqref{Llld}
		\ENDFOR
		\ENDFOR
		\STATE$z_{u}\leftarrow \textup{the concatenation of}\ z_{u,1},z_{u,2},\cdots ,z_{u,k}$
	\end{algorithmic}
\end{algorithm}
\subsubsection{Factor-aware Aggregation}
It is apparent that $e_{u,k}$, the initial node embedding for $k$-th channel, only uses attribute information, with the graph topological structure underutilized. To transform the initial embedding into a high-level embedding, the graph structure information from neighborhoods is required, i.e., to construct $z_{u,k}$ from both $z_{u,k}$ and $\left \{  z_{v,k}:(u,v)\in G \right \}$. However, equally aggregating  all neighbors of node $x_{u}$ may result in excessive interaction with irrelevant information and is not able to reflect the properties of the real-world graph. This, in turn, can lead to incorrect link prediction results.

Therefore, we cannot neglect the diversity of underlying latent factors. The key challenge lies in identifying the neighbors that are connected with the node $u$ due to the latent factor $k$. To tackle this challenge, we propose the factor-aware aggregation strategy to identify and assign neighbors to corresponding channels, and then
perform specific aggregation of neighbors in each channel, inspired by neighborhood routing mechanism \cite{Ma2019}. 
We hypothesize that the more similar the node $u$ and its neighbor $v$ are in the $k$-th channel, the more likely the factor $k$ is to be the reason why node $u$ connects neighbor $v$, which is exemplified by a series of previous studies \cite{Tang2015,10.1145/2939672.2939754,Liu2022}. 
Under the assumption, we compute the probability $p_{v,k}$ that neighbor $v$ should be assigned to $k$-th channel:
\begin{equation}
	p_{v,k}^{(t)}=\frac{exp(e_{v,k}^{\top}z_{u,k}^{(t)})}
	{ \sum _{k=1}^{K}exp(e_{v,k}^{\top}z_{u,k}^{(t)})}  \label{probability},
\end{equation}
where iteration $t=1,2,\cdots,T$, $p_{v,k}\geq 0$ and $\sum _{k=1}^{K} p_{v,k}=1$.
The probability $p_{v,k}$ is intended to perform channel-specific aggregation to update the embedding vector of the $k$-th channel of node $u$:
\begin{equation}
	z_{u,k}^{(t)}=  e_{u,k}+\sum _{v:(u,v)\in G} p_{v,k}^{(t-1)} e_{v,k}
	\label{aggregation}.
\end{equation}
The updated representation is also normalized as $z_{u,k}^{(t)}= {z_{u,k}^{(t)}} / {|| z_{u,k}^{(t)} ||_{2}} $ to be consistent with the initial vector. By performing the factor-aware aggregation (Eq.\eqref{probability}, Eq.\eqref{aggregation}) on all $N$ node, the node embedding matrix for $k$-th channel is obtained: $Z_{k}=[z_{1,k}, \cdots, z_{N,k}]$, $Z_{k} \in \mathbb{R}^{N \times d}$.
\subsubsection{Concatenation}
The final disentangled embedding vector for node $u$ is the concatenation of total $K$ channels, i.e., $z_{u}=concatenate(z_{u,1}, \cdots, z_{u,K})$. By performing the above three operations on all $N$ nodes, the node embedding matrix is $Z=[z_{1},\cdots,z_{N}]$, $Z \in \mathbb{R}^{N \times Kd}$.
\begin{figure}[t!]
	\centering
	\includegraphics[width=1\linewidth]{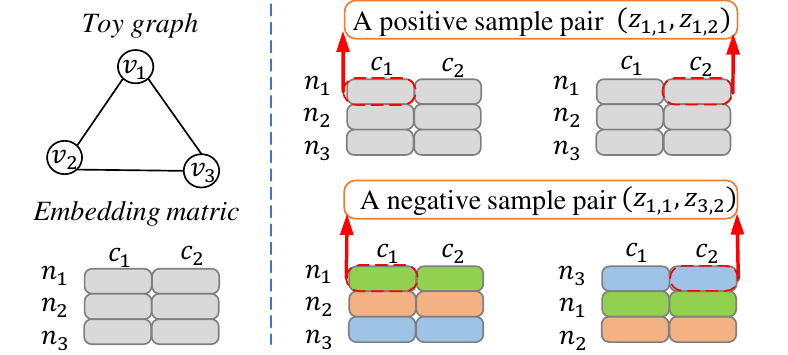}
	\caption{An illustration for positive and negative sampling pairs. The toy graph has 3 nodes $V=\left \{ v_{1},v_{2},v_{3} \right \}$. The embedding matrix is generated by the disentangled graph encoder. For the sketch map of matrix, each row corresponds to  a node, denoted as $n_{i}$ with $i=1,2,3$ , and each column corresponds to a channel, denoted as $c_{j}$ with $j=1,2$. Each square block of the view represents an embedding vector, denoted as $z_{i,j}$, which indicates the $i$-th node embedding vector of $j$-th channel. For example, the positive sample pair is $(z_{1,1}, z_{1,2})$ and the negative sample pair is $(z_{1,1}, z_{3,2})$.}
	\label{sample}
	\vspace{-1em}
\end{figure}
\subsection{Mutual Information Regularization}
Intuitively, the representations of each channel should only be sensitive to specific semantics and not be influenced by other channels. Thus, we encourage independence by minimizing the mutual information (MI) between each of the two channels.
The definition of MI between two different channels of node $u$ is as follows:
\begin{equation}
	I(z_{u,k};z_{u,m})=\mathbb{E}_{p(z_{u,k},z_{u,m})}\left [ \textup{log} \frac{p(z_{u,k},z_{u,m})}{p(z_{u,k})p(z_{u,m})} \right ]
	\label{mi},	
\end{equation}
where $p(z_{u,k},z_{u,m})$ is the joint distribution with $k\neq m$ and $k,m \in\left \{ 1, \cdots, K \right \}$, $p(z_{u,k})$ and $p(z_{u,m})$ represent the respective marginal distributions. Specifically, since it is intractable to directly calculate the exact value of MI in high-dimensional space, we approximate it with  an upper bound of MI: contrastive log-ratio upper bound (CLUB) \cite{Cheng2020}, which has the benefit of capturing the prominent characteristic of graph data. The CLUB-estimator between two channels for node $u$ is defined as follows:
\begin{align}
I_{\textup{CLUB}}(z_{u,k};z&_{u,m})
	=\mathbb{E}_{p(z_{u,k},z_{u,m})}\left [ \textup{log}\, q_{\phi }(z_{u,m}\mid z_{u,k}) \right ] \notag \\
	& -\mathbb{E}_{p(z_{u,k})} \mathbb{E}_{p(z_{u,m})}\left [ \textup{log}\, q_{\phi }(z_{u,m}\mid z_{u,k}) \right ]
	\label{Iclub},
\end{align}
where $q_{\phi }(z_{u,m}\mid z_{u,k})$ is the variational approximation of the unknown conditional probability distribution $p(z_{u,m}\mid z_{u,k})$ with the learnable parameter $\phi$.

The unbiased estimation of variational CLUB between channel $m$ and channel $k$ for all $N$ nodes is as follows:
\begin{equation}
	\mathcal{L}_{mi(k,m)}\!=\!\frac{1}{N}\!\sum_{i=1}^{N}\left [\textup{log}\,q_{\phi }(z_{i,m}\mid z_{i,k})\!-\!q_{\phi }(z_{{i}',m}\mid z_{i,k})  \right ] ,
	\label{Lmi}		
\end{equation}
where $\left (z_{i,k},  z_{i,m} \right ) $ is the positive sample pairs and $ \left (z_{i,k},  z_{{i}',m} \right )$ is negative sample pairs.
\begin{figure}[t!]
	\centering
	\includegraphics[width=1\linewidth]{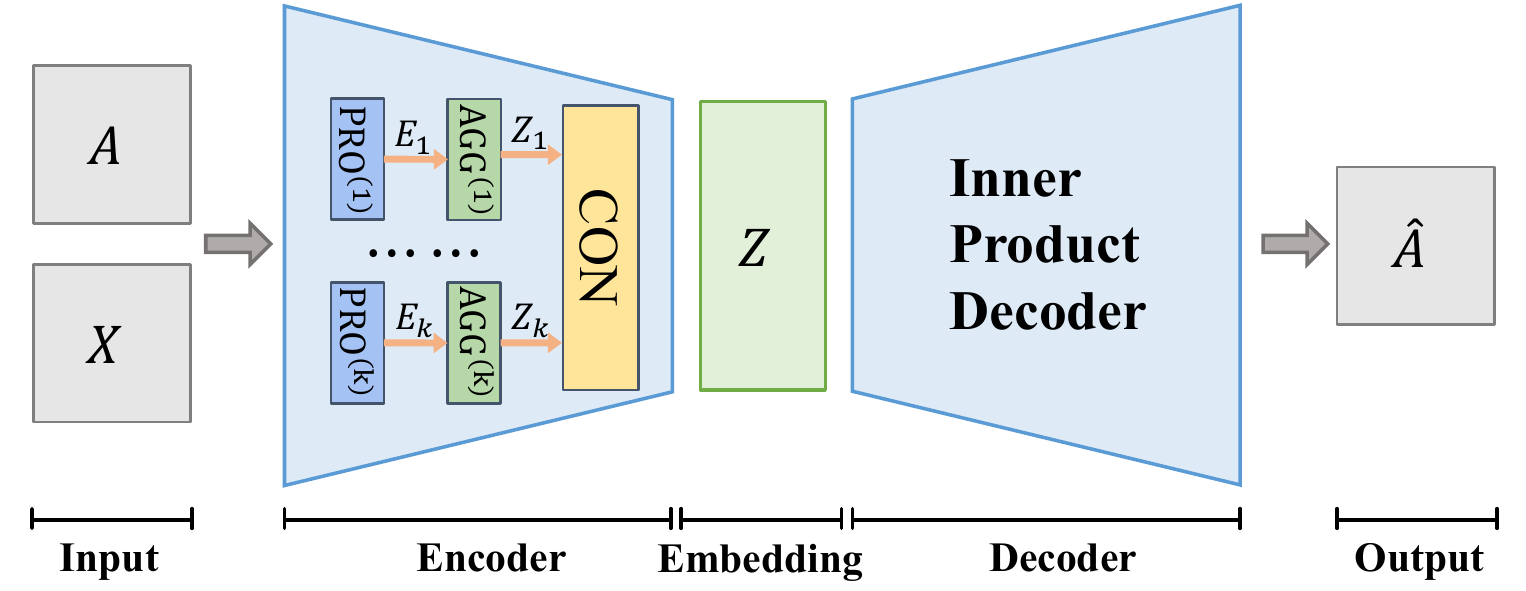}
	\caption{Network architecture. We optimize the entire network by Eq.\eqref{L0} for GDAE or Eq.\eqref{L1} for VGDAE.}
	\label{fig:network}
\end{figure}

We compute the estimation $\mathcal{L}_{mi(k,m)}$ via the contrastive loss between positive and negative sample pairs, where the latter is acquired by keeping the order of the $n$ nodes of channel $k$ and shuffling the $n$ samples of channel $m$. We denote the positive sample set for $k$-th channel as $S_{p}=\left \{ \left (z_{i,k},  z_{i,m} \right )  \right \}_{n}$, the negative sample set for $k$-th channel as $S_{n}=\left \{ \left (z_{i,k},  z_{{i}',m} \right )  \right \}_{n}$ with $k,m=1,2,\cdots, K$ and $k\neq  m$. An illustration for positive and negative sampling pairs is shown in Fig. \ref{sample}.
Moreover, we approximate the unknown conditional probability distribution $q_{\phi }(z_{u,m}\mid z_{u,k})$ with gaussian distribution $\mathcal{N}(z_{u,m} \mid \mu (z_{u,k}),\sigma^{2}(z_{u,k})\cdot \textup{I} )$. 

The mean value $\mu (\cdot)$ and variance $\sigma^{2}(\cdot)$ are computed by the variational approximation network, which consists of two linear layers and a sigmoid layer. 
At the same time, we update the parameter $\phi$ of the variational approximation network, termed $\phi$-VAN, by minimizing the the Kullback-Leibler (KL) divergence between $p(z_{i,k}, z_{i,m})$ and $q_{\phi }(z_{i,k}, z_{i,m})$. In practice, the log-likelihood loss function for optimization is:
\begin{equation}
	\mathcal{L}_{lld}(\phi )=-\frac{1}{N}\sum_{i=1}^{N}\textup{log}\,q_{\phi }(z_{i,m}\mid z_{i,k})
	\label{Llld}.
\end{equation}
\subsection{Network Architecture}	
In this subsection, we describe the overall network architecture of DGAE and VDGAE for link prediction. The network structure is shown in Fig. \ref{fig:network}.
\begin{algorithm}[tb]
	\caption{Link Prediction of GDAE/VDGAE}
	\label{alg:alg}
	\textbf{Input}: Graph $G=(V,E,X)$\\
	\textbf{Output}: $\theta $ of the disentangled graph encoder \par
	\begin{algorithmic}[1] 
		\STATE Randomly initialize the parameters $\theta $. 	
		\WHILE {Not Converged}
		\STATE Calculate node disentangled embedding matrix $Z$ by Eq.\eqref{encoder}.
		\STATE Reconstruct the adjacency matrix $\hat{A}$ by Eq.\eqref{decoder}.
		\STATE Calculate the reconstruction loss by Eq.\eqref{L_rec} for GDAE or Eq.\eqref{L_elbo} for VGDAE.
		\STATE Calculate the MI regularization term $\mathcal{L}_{mi}$ by Eq.\eqref{total_Lmi}.
		\STATE Update $\theta $ by Adam to minimize Eq.\eqref{L0} for GDAE or Eq.\eqref{L1} for VGDAE.
		\ENDWHILE						
		\STATE \textbf{return} $\theta $
	\end{algorithmic}
\end{algorithm}
\subsubsection{Disentangled Graph Encoder}
In order to learn the disentangled node embeddings, we design a disentangled graph encoder, including three layers. The input is a graph $G=(V,E,X)$, with the adjacency matrix $A$. The feature \textbf{pro}ject layer, termed as $  PRO_{\theta_{k}}^{(k)}(\cdot ) $ for $k$-th channel, is used to project the node features into different channels to learn node initial embedding matrix $E_{k}$. The factor-aware \textbf{agg}regation layer, denoted as $AGG^{(k)}(\cdot )$ for $k$-th channel, is used to identify the latent factors and aggregate related neighbors corresponding to each channel to learn $Z_{k}$. Finally, the embedding matrix $Z_{k}$ for total $K$ channel is concatenated to generate the learned node embedding $Z$ and the \textbf{con}catenation layer is denoted as $CON(\cdot)$.

We then consider structure of the three-layer disentangled graph encoder as follows:
\begin{align}
	E_{k}=&PRO_{\theta_{k}}^{(k)}(X,A)  \label{encoder} \\ \notag 
	Z_{k}^{t}=&AGG^{(k)}(E_{k},\ Z_{k}^{(t-1)} )  \\   \notag
	Z=&CON(Z_{1}, Z_{2}, \cdots, Z_{K})    
\end{align}

Furthermore, we introduce MI regularization to restrict the dependence among different channels. 
More details about disentangled graph encoder are shown in Algorithm \ref{alg:algorithm}.

\subsubsection{Inner Product Decoder}
In the following, we consider an inner product decoder for link prediction, with the output of the encoder $z_{u}$ and the adjacency matrix $A \in \mathbb{R}^{N \times N}$:
\begin{align}
&p(\hat{A} \mid Z)=\prod_{i=1}^{N}\prod_{j=1}^{N}p(\hat{A}_{ij} \mid z_{i},z_{j}), \label{decoder}  \\ 
&\textup{with},\,p(\hat{A}_{ij} =1\mid z_{i},z_{j})=\sigma (z_{i}^{\top} z_{j}) \notag
\end{align}
where $\sigma (\cdot )$ is the logistic sigmoid function.

\subsubsection{Optimization} \label{Optimization}
As outlined before, DGAE and VDGAE are variants of graph auto-encoder (GAE) and variational auto-encoder (VGAE) \cite{Kipf2016}, respectively. The reconstruction error for GAE is:
\begin{equation}
	\mathcal{L}_{rec}=\mathbb{E}_{q(Z\mid (X,A))} \left [ \textup{log} \, p(\hat{A}\mid Z) \right ]
	\label{L_rec},
\end{equation}
with $\hat{A}=\sigma (ZZ^{\top})$. Based on the GAE model, VGAE is combined with a negative KL-divergence regularization term to encourage the approximate posterior $q(Z|X,A)$ to be close to the prior $p(Z)$. We assume the prior distribution is the Gaussian distribution, i.e., $p(Z)=\prod_{i}\mathcal{N}(z_{i}|0,I)$. The variational evidence lower bound (ELBO) for VGAE is as follows:
\begin{equation}
	\mathcal{L}_{ELBO}=
	\mathcal{L}_{rec}-KL\left [ \left ( q(Z\mid X,A)\parallel p(Z) \right)\right ]
	\label{L_elbo}.
\end{equation} 
For the MI loss, $\mathcal{L}_{mi(k,m)}$ between every two latent factors requires to be computed, so a total of $K(K-1)/2$ times are calculated for $K$ latent factors due to the symmetry of MI. We calculate the total MI loss function over all nodes as follows:
\begin{equation}
	\mathcal{L}_{mi}=\frac{1}{2}\sum_{k=1}^{K}\sum_{m=1,m\neq k}^{K}\mathcal{L}_{mi(k,m)}	
	\label{total_Lmi}.
\end{equation} 
In general, the final objective function for DGAE:
\begin{equation}
	\mathcal{L}_{0}=\mathcal{L}_{rec}+\lambda _{mi}\mathcal{L}_{mi}	
	\label{L0}.
\end{equation}
The final objective function for VDGAE is:
\begin{equation}
	\mathcal{L}_{1}=\mathcal{L}_{ELBO}+\lambda _{mi}\mathcal{L}_{mi}	
	\label{L1}.
\end{equation}
The weight coefficient $\lambda _{mi} \geq 0$ is employed to control the influence of MI regularization term. In subsequent experiments, a sensitivity analysis will be conducted to evaluate the impact of this parameter on the performance of our proposed methods.

The overall optimization process of DGAE and VDGAE for link prediction is shown in Algorithm \ref{alg:alg}.
\begin{table*}
	\centering
	\caption{Dataset statistics}
	\setlength\tabcolsep{2.0pt}
	\begin{tabular}{lcccccccccccccccccc}
		\toprule
		Datasets  &USAir  &NS   &PB &Yeast &C.ele &Power &Router &E.coli &Cora  &Citeseer &Pubmed &Chameleon &Squirrel &Cornell &Texas &Wisconsin &Computers &Photo  \\
		\midrule
		Nodes        &332  &1589 &1222  &2375  &297  &4941 &5022 &1805  &2708  &3327  &19717 &2277  &5201    &183  &183  &251 &13752  &7650  \\
		Links   	 &2126 &2742 &16714 &11693 &2148 &6594 &6258 &15660 &5278  &4552  &44324 &36101 &217073  &298  &325  &515 &491722 &238162  \\
		Features     &No   &No   &No    &No    &No   &No   &No   &No    &1433  &3703  &500   &2325  &2089    &1703 &1703 &1703 &767    &745   \\      	
		\bottomrule
	\end{tabular}
	\label{tab:Dataset w attr}
\end{table*}
\subsection{Complexity Analysis}
The parameters of DGAE and VDGAE are quite small. The total number of parameters is $\mathcal{O}(K(N*d+d))$, i.e. $\mathcal{O}(N)$, where $K$ is the number of channels, $N$ is the number of nodes and $d$ is the dimensionality of the embedding vector for each channel. Since the proposed models involve the reconstruction of the adjacency matrix, the computation complexity is $\mathcal{O}(N^2)$, where $N$ denotes the number of nodes in the graphs.

\section{Experiment}

In this section, we demonstrate the ability of the VDGAE and DGAE models on link prediction against various baselines, and conduct a series of experimental analysis.
\subsection{Experimental Setup}
\subsubsection{Datasets}
Two groups of benchmarks are used to verify the performance of the proposed model on link prediction. One group of datasets are eight datasets without node attributes, which includes USAir \cite{USAir}, NS \cite{newman2006finding}, PB \cite{Ackland2005}, Yeast \cite{VonMering2002}, C.ele \cite{Watts1998}, Power \cite{Watts1998}, Router \cite{Spring2002}, E.coli \cite{Zhang2018a}. The other group of datasets is ten graphs with node attributes. The datasets with node attributes are: 
\begin{itemize}
	\item{Citation network datasets: Cora, Citeseer, and Pubmed \cite{Yang2016}, where these datasets are the most widely used benchmarks for link prediction. Nodes represent documents and edges represent citation links in these three graphs.}
	\item{WikipediaNetwork datasets: Chameleon, Squirrel \cite{DBLP:conf/iclr/VelickovicCCRLB18}, where nodes represent web pages and edges represent hyperlinks between nodes. These two graphs are used as graphs with certain degrees of heterophily \cite{Zhu2020}.}
	\item{Webpage datasets: Cornell, Texas, Wisconsin \cite{Pei2020Geom-GCN:}, where nodes represent web pages and edges represent hyperlinks between nodes.}
	\item{Amazon datasets: Computers, Photo \cite{shchur2018pitfalls}, where nodes represent goods and edges represent that two goods are frequently bought together. These two graphs are larger graphs with more nodes and edges and hence more challenging compared
	to the citation datasets.}
\end{itemize}
The statistics of the above datasets are summarized in Table \ref{tab:Dataset w attr}.
We preprocess the graph in these datasets via the library PyG (PyTorch Geometric) \cite{Fey/Lenssen/2019}. To avoid the link leak problem, duplicate edges are removed for all datasets when converting the graph to an undirected one.


\subsubsection{Baselines}
We compare the proposed methods with two groups of baselines based on heuristic methods and embedding methods. The first groups of baselines are traditional heuristic methods for experiments on datasets without node attribute information. The heuristic baselines include: 
\begin{itemize}
	\item{Four path-based methods: Adamic-Adar (AA) \cite{Adamic2003}, preferential attachment (PA) \cite{Barabasi1999}, resource allocation (RA) \cite{Zhou2009}, and Katz \cite{Katz1953}.}
	\item{Two neighbor-based methods: Common neighbors (CN) \cite{Newman2001}, Jaccard\cite{Jaccard1902}.}
	\item{One walk-based methods: SimRank (SR) \cite{Jeh2002}.}
	\item{Two based on enclosing subgraphs methods: Weisfeiler-Lehman graph kernel (WLK) \cite{Shervashidze2011} and WLNM\cite{Zhang2017}.}
\end{itemize}

The other groups of baselines based on node embeddings for experiments on datasets with node attribute information include:
\begin{itemize}
	\item{Several variants of graph auto-encoder: graph auto-encoder (GAE) \cite{Kipf2016}, variational graph auto-encoder (VGAE) \cite{Kipf2016}, hyperspherical variational graph auto-encoder (s-VAE) \cite{DBLP:conf/uai/DavidsonFCKT18}, adversarially regularized variational graph auto-encoder (ARVGE) \cite{DBLP:conf/ijcai/PanHLJYZ18}, Graphite-VAE \cite{Grover2019}, graph normalized auto-encoder (GNAE) \cite{10.1145/3459637.3482215}, variational graph normalized auto-encoder (VGNAE) \cite{10.1145/3459637.3482215}.}
	\item{Classical unsupervised graph representation learning methods: DeepWalk \cite{10.1145/2623330.2623732}, Node2vec \cite{10.1145/2939672.2939754} and Spectral Clustering (SC) \cite{tang2011leveraging}.}
	\item{Graph InfoClust (GIC) \cite{DBLP:conf/pakdd/MavromatisK21} and deep graph infomax (DGI) \cite{DBLP:conf/iclr/VelickovicFHLBH19}, which are the two state-of-the-art methods involved graph-level information.}
\end{itemize}
\subsubsection{Data Splitting and Evaluation Metrics}
For the heuristic baselines, we follow the experiments set as \cite{Zhang2018}, where 90\% of existing links are observed during training. For the baselines based on embedding methods, we perform link prediction tasks using the same dataset splits as \cite{Kipf2016}: for each dataset, the validation and test sets contain 5\% and 10\% of citation links, respectively. Meanwhile, the performance is reported on the standard link prediction metrics: area under the ROC curve (AUC) and average precision (AP) scores. We optimize hyperparameters on the validation set and report the average of the metrics for five runs on the test set.
\begin{table*}[ht]
	\centering 
	\caption{Results of link prediction for heuristic baselines on datasets without node attributes, in terms of AUC.}	
	\setlength\tabcolsep{7pt}
	\begin{tabular}{lcccccccc}
		\toprule
		Dataset   &USAir &NS &PB &Yeast &C.ele &Power &Router &E.coli  \\                    
		\midrule
		AA \cite{Adamic2003} &95.06 ± 1.03 &94.45 ± 0.93 &92.36 ± 0.34 &89.43 ± 0.62 &86.95 ± 1.40 &58.79 ± 0.88 &56.43 ± 0.51 &95.36 ± 0.34 \\
		CN \cite{Newman2001} &93.80 ± 1.22 &94.42 ± 0.95 &92.04 ± 0.35 &89.37 ± 0.61 &85.13 ± 1.61 &58.80 ± 0.88 &56.43 ± 0.52 &93.71 ± 0.39\\
		PA \cite{Barabasi1999} &88.84 ± 1.45 &68.65 ± 2.03 &90.14 ± 0.45 &82.20 ± 1.02 &74.79 ± 2.04 &44.33 ± 1.02 &47.58 ± 1.47 &91.82 ± 0.58\\
		RA \cite{Zhou2009} &95.77 ± 0.92 &94.45 ± 0.93 &92.46 ± 0.37 &89.45 ± 0.62 &87.49 ± 1.41 &58.79 ± 0.88 &56.43 ± 0.51 &95.95 ± 0.35\\
		Jaccard \cite{Jaccard1902} &89.79 ± 1.61 &94.43 ± 0.93 &87.41 ± 0.39 &89.32 ± 0.60 &80.19 ± 1.64 &58.79 ± 0.88 &56.40 ± 0.52 &81.31 ± 0.61\\
		Katz \cite{Katz1953}    &92.88 ± 1.42 &94.85 ± 1.10 &92.92 ± 0.35 &92.24 ± 0.61 &86.34 ± 1.89 &65.39 ± 1.59 &38.62 ± 1.35 &93.50 ± 0.44\\
		SR \cite{Jeh2002} &78.89 ± 2.31 &94.79 ± 1.08 &77.08 ± 0.80 &91.49 ± 0.57 &77.07 ± 2.00 &76.15 ± 1.06 &37.40 ± 1.27 &62.49 ± 1.43\\
		WLK \cite{Shervashidze2011}  &96.63 ± 0.73 &98.57 ± 0.51 &\textbf{93.83 ± 0.59} &95.86 ± 0.54 &89.72 ± 1.67 &82.41 ± 3.43 &87.42 ± 2.08 &96.94 ± 0.29\\
		WLNM \cite{Zhang2017} &95.95 ± 1.10 &98.61 ± 0.49 &93.49 ± 0.47 &95.62 ± 0.52 &86.18 ± 1.72 &84.76 ± 0.98 &94.41 ± 0.88 &\textbf{97.21 ± 0.27}\\
		\hline
		DGAE (Ours)   &94.49 ± 1.44 &99.79 ± 0.22 &91.09 ± 0.77 &\textbf{98.44 ± 0.18} &\textbf{91.07 ± 1.41} &98.78 ± 0.13 &97.23 ± 0.20 &92.92 ± 0.55\\
		VDGAE (Ours)  &\textbf{96.73 ± 1.18} &\textbf{99.84 ± 0.08} &90.06 ± 0.85 &98.05 ± 0.24 &89.05 ± 1.06 &\textbf{99.76 ± 0.08} &\textbf{97.29 ± 0.81} &91.99 ± 0.47\\
		\bottomrule
	\end{tabular}
	\label{tab:results_1}
\end{table*}
\begin{table*}[ht]
	\centering 
	\caption{Results of link prediction on Cora, Citeseer, and Pubmed, in terms of AUC and AP.}
	\setlength\tabcolsep{10pt}
	\begin{tabular}{lcccccc}
		\toprule
		\multirow{2}{*}{Dataset}   &\multicolumn{2}{c}{Cora} 
		&\multicolumn{2}{c}{Citesee}   &\multicolumn{2}{c}{Pubmed}   \\  
		\cmidrule(lr){2-3}  \cmidrule(lr){4-5}  \cmidrule(lr){6-7}  
		&AUC &AP &AUC &AP &AUC &AP   \\                         
		\midrule
		SC \cite{tang2011leveraging}                          &84.60 ± 0.01 &88.50 ± 0.00 &80.50 ± 0.01 &85.00 ± 0.01 &84.20 ± 0.02 &87.80 ± 0.01  \\ 
		DeepWalk  \cite{10.1145/2623330.2623732}              &83.10 ± 0.01 &85.00 ± 0.00 &80.50 ± 0.02 &83.60 ± 0.01 &84.40 ± 0.00 &84.10 ± 0.00  \\ 
		Node2vec  \cite{10.1145/2939672.2939754}              &85.60 ± 0.15 &87.50 ± 0.14 &89.40 ± 0.14 &91.30 ± 0.13 &91.90 ± 0.04 &92.30 ± 0.05  \\ 
		GAE \cite{Kipf2016}                                   &89.54 ± 1.65 &89.75 ± 1.51 &88.73 ± 0.84 &89.07 ± 1.05 &95.69 ± 0.12 &95.76 ± 0.10  \\                     
		VGAE \cite{Kipf2016}                                  &85.22 ± 4.93 &81.02 ± 3.39 &81.02 ± 3.39 &81.90 ± 2.97 &92.91 ± 1.34 &92.79 ± 1.55  \\                       
		s-VAE  \cite{DBLP:conf/uai/DavidsonFCKT18}            &94.10 ± 0.10 &81.90 ± 2.97 &94.70 ± 0.20 &95.20 ± 0.20 &96.00 ± 0.10 &96.00 ± 0.10 \\                           	
		ARVGE \cite{DBLP:conf/ijcai/PanHLJYZ18}               &91.27 ± 0.79 &91.39 ± 0.91 &87.78 ± 1.77 &88.09 ± 1.96 &96.47 ± 0.15 &96.52 ± 0.09  \\    	 
		Graphite-VAE \cite{Grover2019}                        &94.70 ± 0.11 &94.90 ± 0.13 &97.30 ± 0.06 &97.40 ± 0.06 &97.40 ± 0.04 &97.40 ± 0.04  \\ 
		GNAE \cite{10.1145/3459637.3482215}                   &94.08 ± 0.63 &94.40 ± 0.95 &96.90 ± 0.22 &97.19 ± 0.12 &95.41 ± 0.19 &94.86 ± 0.35  \\   
		VGNAE \cite{10.1145/3459637.3482215}                  &89.19 ± 0.67 &89.58 ± 0.75 &95.45 ± 0.55 &95.70 ± 0.52  &89.66 ± 0.40 &89.36 ± 0.25  \\  
		GIC  \cite{DBLP:conf/pakdd/MavromatisK21}             &93.50 ± 0.60 &93.30 ± 0.70 &97.00 ± 0.50 &96.80 ± 0.50 &93.70 ± 0.30 &93.50 ± 0.30  \\ 	
		DGI  \cite{DBLP:conf/iclr/VelickovicFHLBH19}          &89.80 ± 0.80 &89.70 ± 1.00 &95.50 ± 1.00 &95.70 ± 1.00 &91.20 ± 0.60 &92.20 ± 0.50  \\ 		
		\hline
		DGAE (Ours)    &95.80 ± 0.44 &96.07 ± 0.34 &97.23 ± 0.34 &97.46 ± 0.25 &\textbf{97.77 ± 0.12} &\textbf{97.82 ± 0.15} \\ 	
		VDGAE (Ours)    &\textbf{95.90 ± 0.42} &\textbf{96.17 ± 0.29} &\textbf{97.82 ± 0.30} &\textbf{98.03 ± 0.22} &97.00 ± 0.12 &97.07 ± 0.15  \\	
		\bottomrule
	\end{tabular}
	\label{tab:results}
\end{table*}
\subsubsection{Parameter Settings} \label{parameset}
The proposed DGAE and VDGAE are implemented in PyTorch on NVIDIA GeForce RTX 3090 by using the Adam optimizer \cite{DBLP:journals/corr/KingmaB14}. For the link prediction task, the output dimension of DGAE and VDGAE sets as $K\Delta d$. The number of channels $K$ is searched from $\left \{ 2,3,\cdots ,10 \right \}$. The output dimension of each channel is $\Delta d=64$. The iteration of routing is $T=3$ and the iteration of the $\phi$-VAN, i.e., inner loop is $M=5$. Specifically, the learning rate of the proposed methods and the $\phi$-VAN are both tuned in $\left \{ 0.001, 0.002,\cdots ,1.000 \right \}$. The coefficient of mutual information regularization term $\lambda _{mi}$ are selected from $\left \{ 0.01, 0.02,\cdots ,1.00 \right \}$. 
We then tune the hyperparameters using grid search for each hyperparameter combination. Moreover, the parameters are initialized with Xavier for a fair comparison with the proposed models and all baselines.

\subsection{Performance Comparison}
\begin{table*}[ht]
	\centering 
	\caption{Results of link prediction on eight datasets with node attributes, in terms of AUC.}	
	\setlength\tabcolsep{6.3pt}
	\begin{tabular}{lcccccccc}
		\toprule
		Dataset   &Wisconsin &Cornell &Texas &Computers &Photo &Squirrel &Chameleon  &Crocodile   \\                    
		\midrule
		GAE  \cite{Kipf2016}       &68.87 ± 3.84 &73.60 ± 10.90 &75.34 ± 12.97 &86.26 ± 1.52 &87.32 ± 1.20 &94.09 ± 0.09 &97.03 ± 0.35 &95.27 ± 0.16\\
		GNAE \cite{10.1145/3459637.3482215}        &78.17 ± 8.29 &72.93 ± 10.83 &75.14 ± 10.67  &92.17 ± 0.12 &94.64 ± 0.11 &93.22 ± 0.07 &\textbf{98.30 ± 0.09} &94.47 ± 0.11\\
		VGAE  \cite{Kipf2016}      &66.94 ± 8.66 &78.30 ± 4.01 &76.68 ± 5.57 &87.14 ± 0.27 &87.03 ± 0.42 &94.05 ± 0.10 &96.42 ± 0.18  &94.11 ± 0.26\\	
		ARVGE  \cite{DBLP:conf/ijcai/PanHLJYZ18}   &71.10 ± 3.77 &\textbf{78.88 ± 5.01} &76.46 ± 4.68 &82.97 ± 0.67 &81.07 ± 0.52	&93.30 ± 0.17 &92.86 ± 0.27 &94.39 ± 0.16 \\	
		VGNAE  \cite{10.1145/3459637.3482215}         &70.26 ± 1.20 &73.28 ± 5.73  &78.93 ± 3.02   &80.68 ± 0.10 &79.48 ± 0.23  &89.37 ± 0.12 &95.36 ± 0.17 &90.50 ± 0.42\\
		\hline
		DGAE (Ours)      &75.73 ± 5.86 &68.05 ± 12.07 &68.27 ± 12.79 &\textbf{94.12 ± 0.19} &\textbf{95.49 ± 0.15}  &\textbf{96.63 ± 0.04} &97.05 ± 0.10 &\textbf{96.18 ± 0.07}\\
		VDGAE (Ours)      &\textbf{85.03 ± 4.78} &76.08 ± 4.75  &\textbf{81.30 ± 8.49 } &93.29 ± 0.21  &94.68 ± 0.10  &95.78 ± 0.08 &96.85 ± 0.14 &93.90 ± 0.18\\
		\bottomrule
	\end{tabular}
	\label{tab:results_2}
\end{table*}
\subsubsection{Comparison to heuristic methods}
Table \ref{tab:results_1} summarizes the link prediction results for heuristic baselines on datasets without node attributes. Our observations show that the proposed DGAE and VDGAE methods significantly outperform all baselines on USAir, NS, Yeast, C.ele, Power, and Router datasets, with particularly impressive results on the Power dataset. In fact, our methods achieve a prediction accuracy improvement of approximately 15\% on the Power dataset over the WLNM baseline model, which has the highest accuracy among all baselines. 
While our methods perform slightly worse than the best-performing baseline in previous work on the PB and E.coli datasets,  the difference is not significant.
\subsubsection{Comparison to embedding methods} 
Table \ref{tab:results} summarizes the link prediction results for embedding baselines on three citation network datasets. 
The experimental results demonstrate that both DGAE and VDGAE outperform all competitive baselines on three citation network datasets. Specifically, our proposed methods achieve significant improvements over all classical unsupervised methods (e.g., DeepWalk, SC, and Node2vec) on all three datasets. Moreover, it can be seen that the disentangled mechanism provides a substantial boost compared to other variants of VGAE. For instance, VDGAE surpasses Graphite-VAE by 1.2\% on Cora and 0.5\% on Citeseer, where Graphite-VAE is considered as the baselines with the best predictive performance. Additionally, DGAE also outperforms Graphite-VAE by 0.4\% on Pubmed.
It is worth noting that our framework does not directly rely on the adjacency matrix during the aggregation process of each channel, yet still achieves better performance than the baselines. This is because neighborhood information is automatically incorporated with factor-aware aggregation. 

Table \ref{tab:results_2} shows the link prediction results for embedding baselines on eight other datasets with node attributes. It is apparent that our methods perform exceptionally well on six datasets and slightly worse on the remaining two datasets (Cornell, Chameleon) compared to five variants of VGAE. Notably, on the Wisconsin dataset, VDGAE exhibits outstanding performance, surpassing other variants by approximately 15\%. Additionally, the proposed methods achieve excellent results on two large graphs (Computers, Photo), indicating that the node embeddings learned by the disentanglement mechanism have better generalization ability to adapt to more nodes.

Extensive experiments illustrate the strengths of the proposed models--DGAE and VDGAE tend to capture rich underlying information corresponding to latent factors on the graph and learn disentangled node representations, consequently leading to substantial enhancements in the performance of link prediction tasks on many different datasets.
\subsection{Ablation Study}
To quantitatively investigate the effectiveness of the main modules, we conduct ablation studies to analyze the impact of the disentanglement mechanism and independence constraints on DGAE and VDGAE, respectively. 
We compare the proposed methods with the following two variants: (1) $w/o\ MI$: where we remove the MI regularization term, causing the disentangled graph encoder to degenerate into DisenGCN \cite{Ma2019}; (2) $w/o\ disent$: where we deprecate the disentangled mechanism to explore the impact of overall disentanglement module. In this case, the node embeddings obtained by feature projection serve as the input of the decoder. Leaving entanglement unexplored, DGAE actually degrades into the linear GAE model, and VDGAE degrades into VGAE.

The results of ablation studies are illustrated in Table \ref{tab:Ablation_DGAE} for DGAE and in Table \ref{tab:Ablation_VDGAE} for VDGAE.
We set the parameters for the respective variants according to the experimental settings of DGAE and VDGAE, and report the results in terms of AUC on three citation network datasets. It is evident that the absence of disentanglement causes a significant drop in performance,  with a reduction of around 3\%-4\% for DGAE and 5.2\%-9\% for VDGAE. This demonstrates the critical role of inferring latent factors in improving the performance of the link prediction task. Moreover, our results show that the removal of CLUB results in a performance decline of 0.6\%-1\% for DGAE and 0.2\%-0.9\% for VDGAE, highlighting the importance of the proposed independence constraints. Hence, we can conclude that the disentanglement mechanism and independence constraints are essential for maintaining the strong predictive performance of our methods.
\begin{table}[tp]
	\centering
	\caption{Results of ablation study for DGAE.}
	\begin{tabular}{lccccccc}
		\toprule
		Dataset   &Cora &Citesee   &Pubmed  \\
		\midrule		
		w/o MI  &94.80 ± 0.65    &96.67 ± 0.66      &96.71 ± 0.08            \\
		w/o disent    &92.36 ± 0.65    &94.06 ± 0.20      &93.89 ± 0.16            \\
		DGAE       	  &95.80 ± 0.44    &97.23 ± 0.34      &97.77 ± 0.12           \\   		
		\bottomrule
	\end{tabular}
	\label{tab:Ablation_DGAE}
\end{table}
\begin{table}[tp]
	\centering
	\caption{Results of ablation study for VDGAE.}
	\begin{tabular}{lccccccc}
		\toprule
		Dataset   &Cora &Citesee   &Pubmed          \\
		\midrule
		w/o MI       &95.14 ± 0.18   &96.90 ± 0.59    &96.78 ± 0.10      \\
		w/o disent   &86.93 ± 0.49   &90.16 ± 1.10    &91.87 ± 0.28      \\		
		VDGAE         &95.90 ± 0.42   &97.82 ± 0.30    &97.00 ± 0.12      \\   		
		\bottomrule
	\end{tabular}
	\label{tab:Ablation_VDGAE}
\end{table}

\begin{figure*}[h]
	\centering
	\includegraphics[width=1\linewidth]{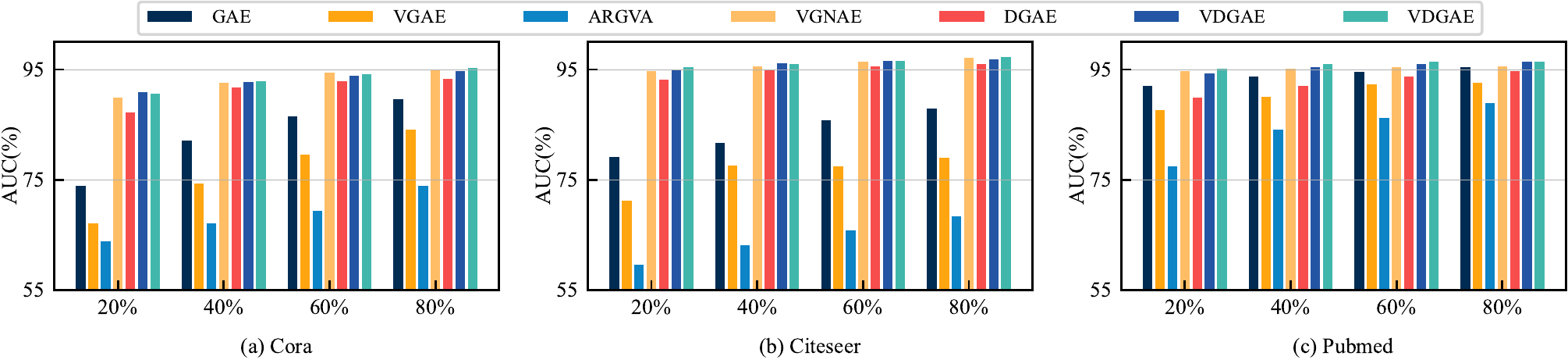}
	\caption{Results of link prediction on three citation network datasets, where the training set varies at 20\%$|E|$, 40\%$|E|$, 60\%$|E|$ and 80\%$|E|$ of the total number of edges $|E|$, in terms of AUC.}
	\label{fig:split}
\end{figure*}
\begin{figure}[t]
	\centering
	\subfloat[Impact of channels]{\includegraphics[width=0.49\linewidth]{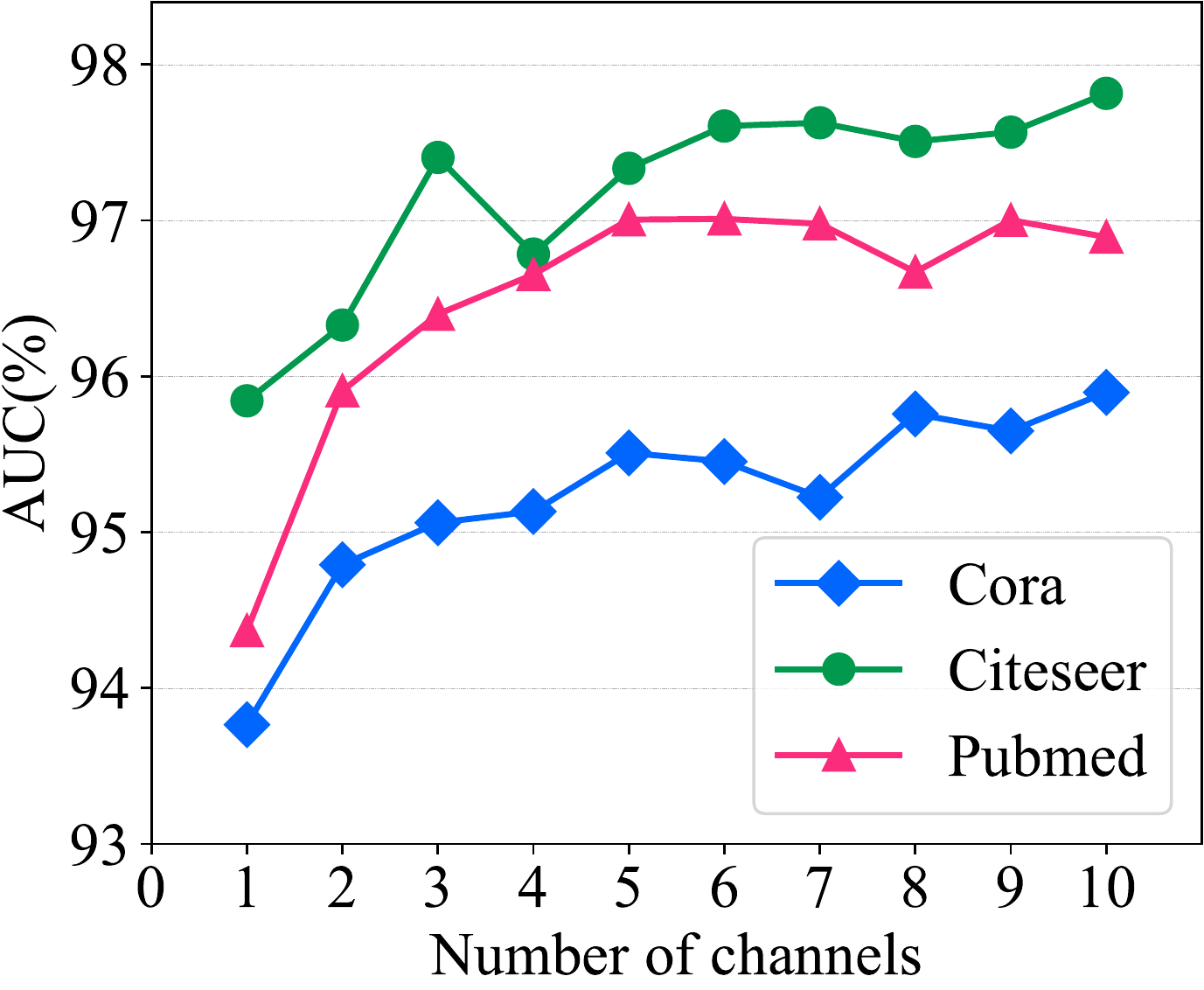}}
	\label{fig:channels}			
	\subfloat[Impact of $\lambda _{mi}$]{\includegraphics[width=0.49\linewidth]{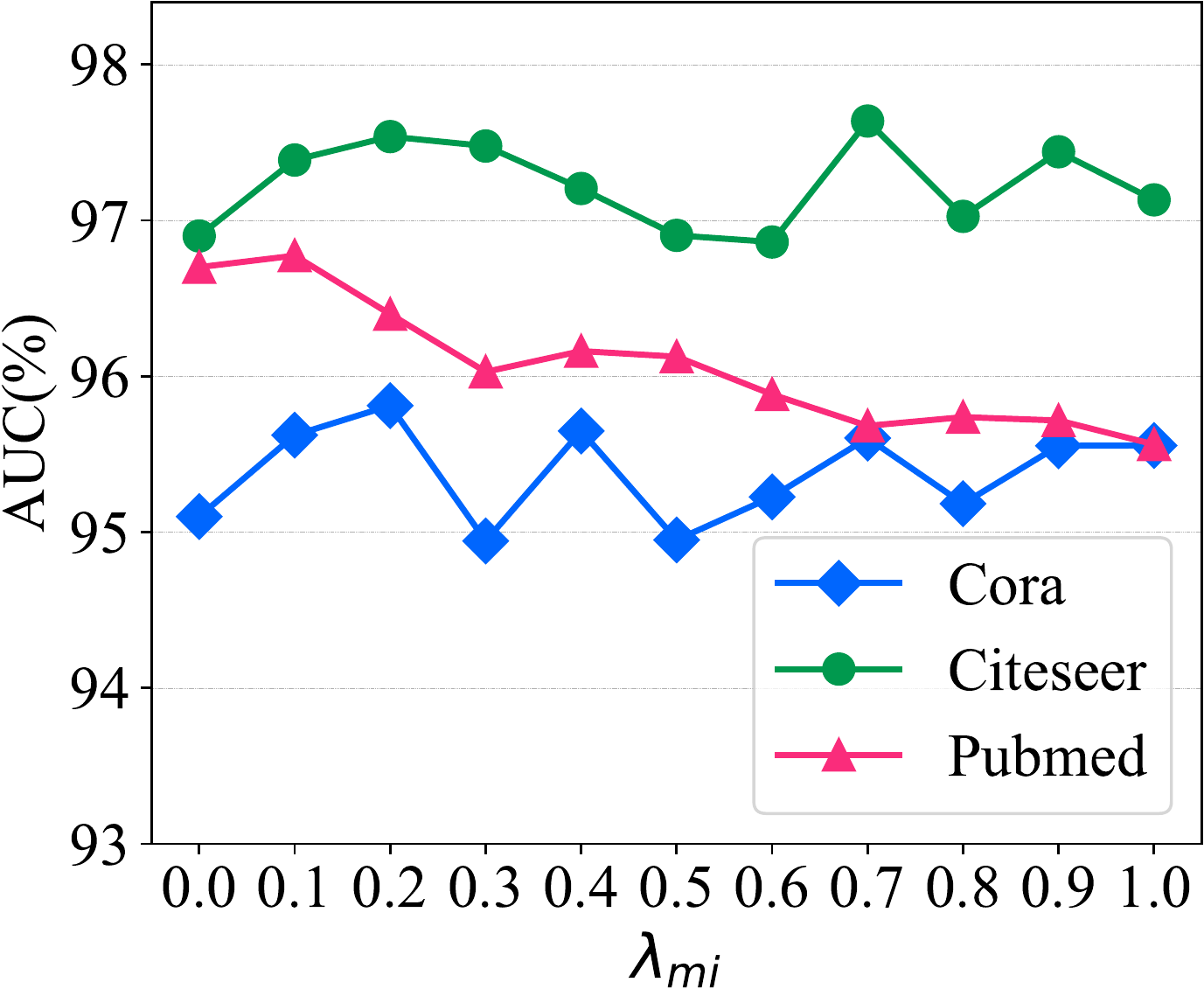}}
	\label{fig:lambda}		
	\subfloat[Loss]{\includegraphics[width=0.49\linewidth]{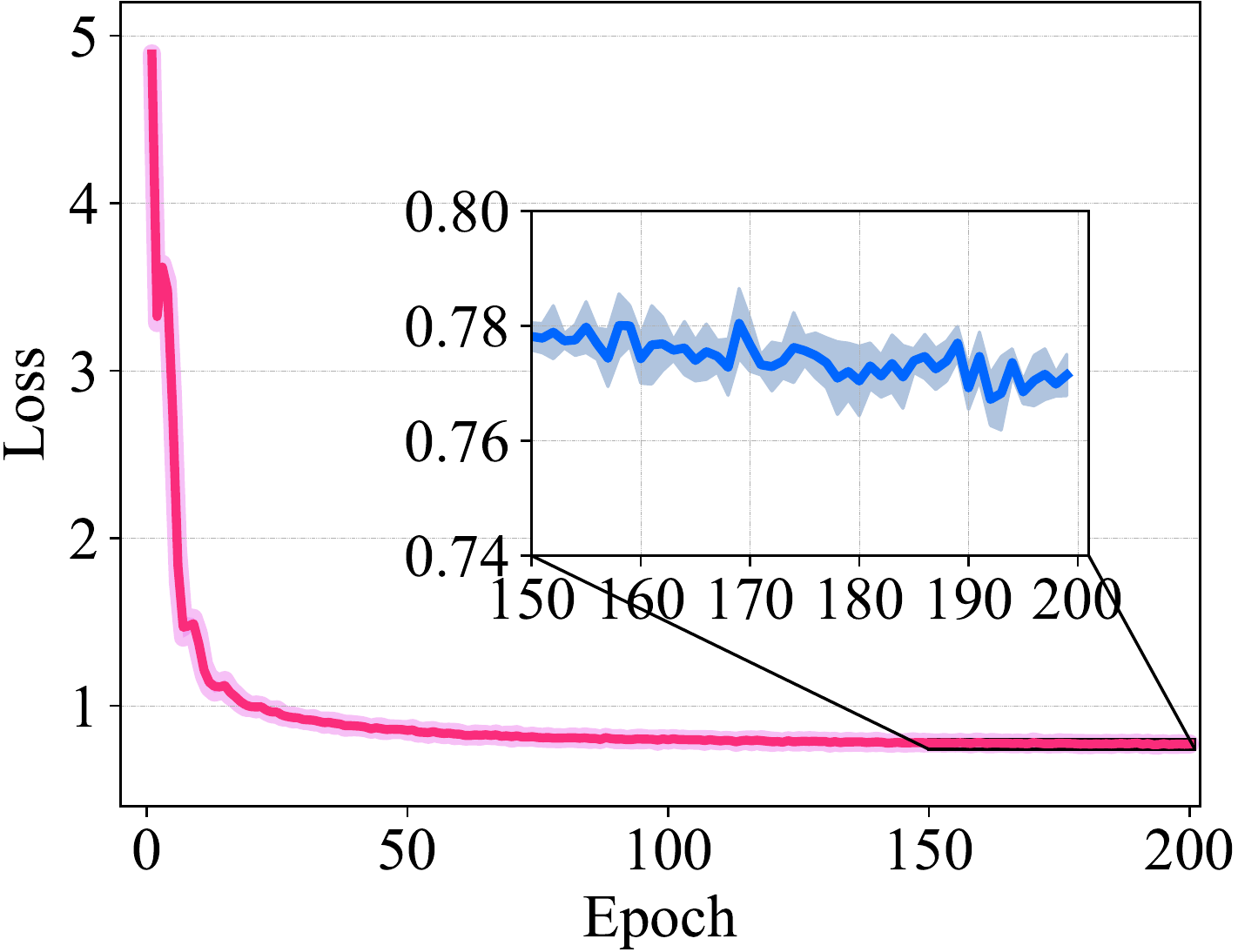}}
	\label{fig:loss}			
	\subfloat[AUC]{\includegraphics[width=0.49\linewidth]{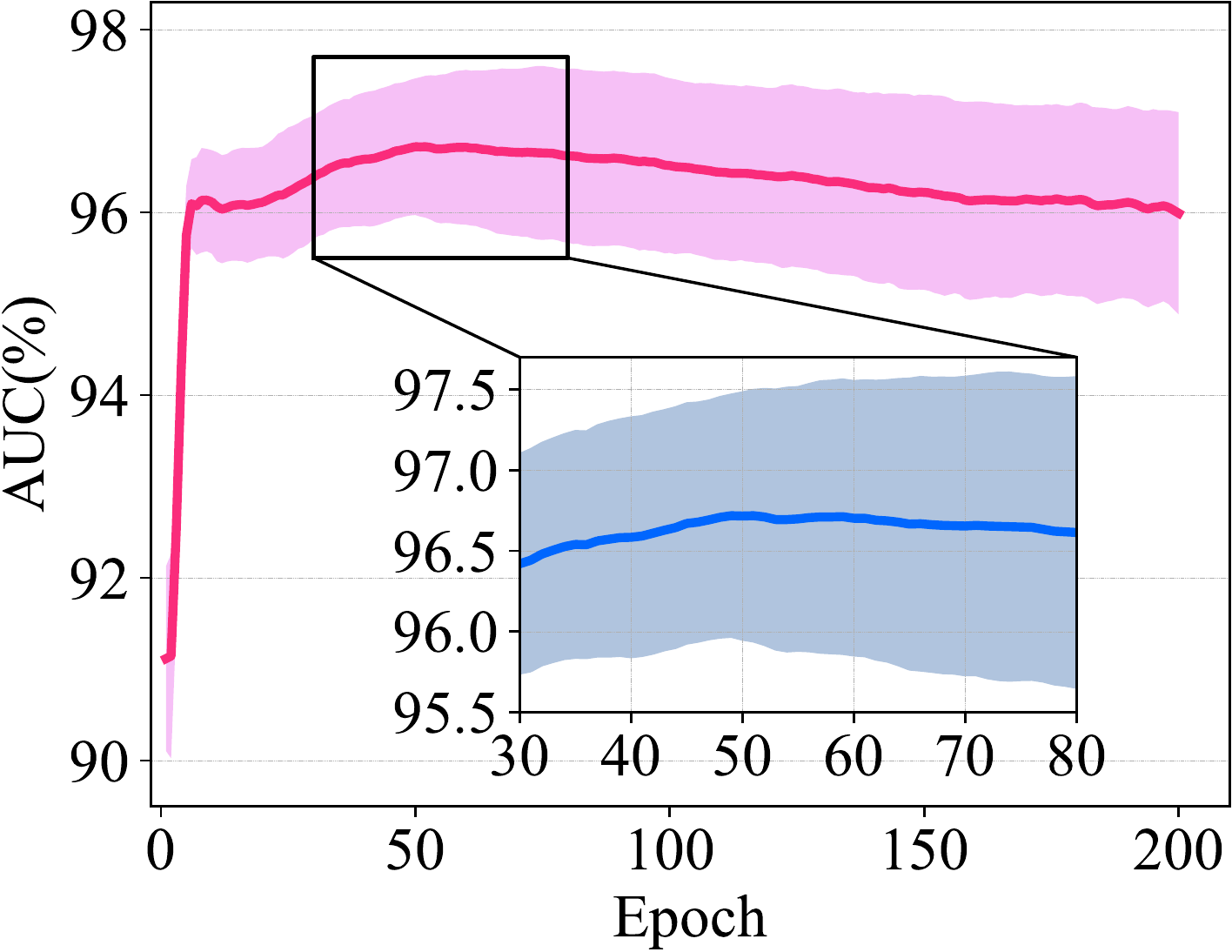}}
	\label{fig:auc}	
	\caption{Hyperparameter sensitivity and convergence analysis of VDGAE on Cora, Citeseer, and Pubmed in terms of AUC.}
	\label{fig:hyp_sens}
\end{figure}
\subsection{Sensitivity Analysis}
\subsubsection{Hyperparameter Sensitivity}
In this part, we investigate the effect of two key hyperparameters on the proposed models: the number of channels $K$, and the coefficient of mutual information regularization term $\lambda _{mi}$. Fig. \ref{fig:hyp_sens}(a) and Fig. \ref{fig:hyp_sens}(b) show the experimental results for hyperparameter sensitivity of VDGAE on three citation network datasets in terms of AUC. 

The results from Fig. \ref{fig:hyp_sens}(a) reveal a similar tendency on three datasets: as the number of channels varies from 2 to 10, the AUC value gradually increases. However, increasing $K$ beyond 7 no longer renders a clear marginal rise, but instead increases the runtime by a large factor of 10×. Therefore, we consider the first turning point from rising to plateau as the final parameter $K$ applied in training, i.e., for Cora, $K=5$; for Citeseer, $K=6$; for Pubmed, $K=5$. Moreover, when $K=1$, the encoder of VDGAE degrades into GCN \cite{kipf2017semisupervised}. Fig. \ref{fig:hyp_sens}(b) shows how the parameter $\lambda _{mi}$ in Eq.\eqref{L1} affects the performance of VDGAE. We tune $\lambda _{mi}$ on 10 points uniformly obtained between 0 and 1, and check the corresponding results. For $\lambda _{mi}=0$, the VDGAE model degenerates into the first variant of ablation experiments, $w/o\ CLUB$. Furthermore, we observe that 
the parameter $\lambda _{mi}$ is beneficial within the range $\left [ 0.1, 0.3 \right ]$ for Cora, $\left [ 0.6, 0.8 \right ]$ for Citeseer, and $( 0, 0.2] $ for Pubmed.
\subsubsection{Convergence Analysis}
We show the variation of the average loss in Fig. \ref{fig:hyp_sens}(c) and the average AUC in Fig. \ref{fig:hyp_sens}(d) over five runs on the training set of Cora dataset. The plot in Fig. \ref{fig:hyp_sens}(c) clearly shows that VDGAE converges in a small number of iterations, which empirically proves the efficiency of the proposed model. Moreover, the varying AUC curve over multiple runs in Fig. \ref{fig:hyp_sens}(d) demonstrates the stability of our method.
\subsubsection{Dataset Segmentation}
We evaluate the performance of several VGAE-based baselines on three citation network datasets, by training with 20\%$|E|$, 40\%$|E|$, 60\%$|E|$ and 80\%$|E|$ of the total number of edges $|E|$. 
The edges are divided into a validation set of 5\%$|E|$ and the remaining for the test set.
As can be seen in Fig. \ref{fig:split}, DGAE and VDGAE consistently outperform other methods across all divisions. We can observe that our proposed methods perform significantly better when fewer edges are observed compared to the baseline models. The observation suggests that the proposed methods are more prone to generalize to unseen nodes in the graph.

\section{Qualitative Analysis}
We provide qualitative analysis from two aspects, analysis of disentanglement performance and visualization of node embeddings.
\begin{figure*}[t!]
	\centering
	\subfloat[w/o\ disent]{\includegraphics[width=0.24\linewidth]{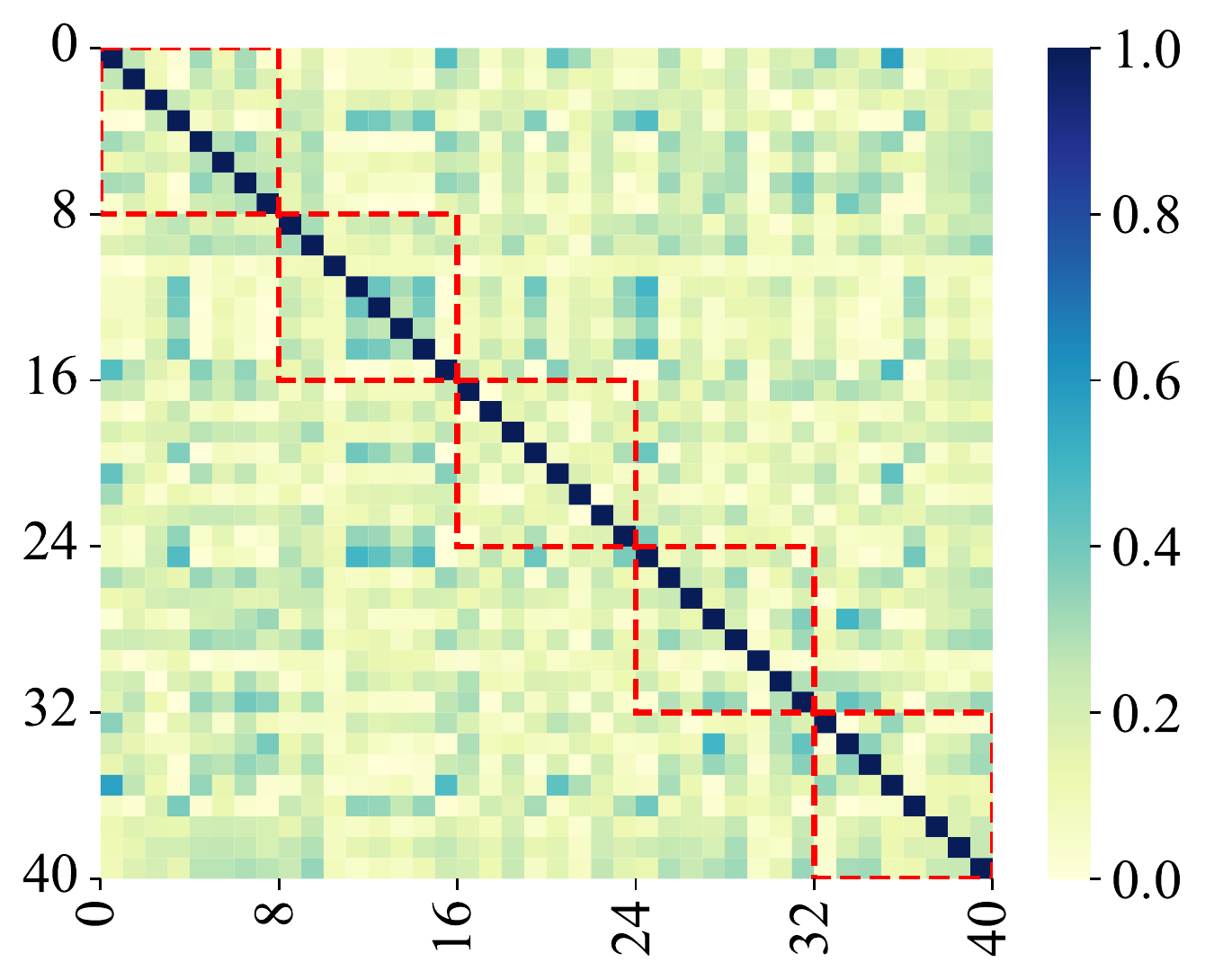}}\label{fig:raw_feat}			
	\hfil
	\subfloat[VGAE]{\includegraphics[width=0.24\linewidth]{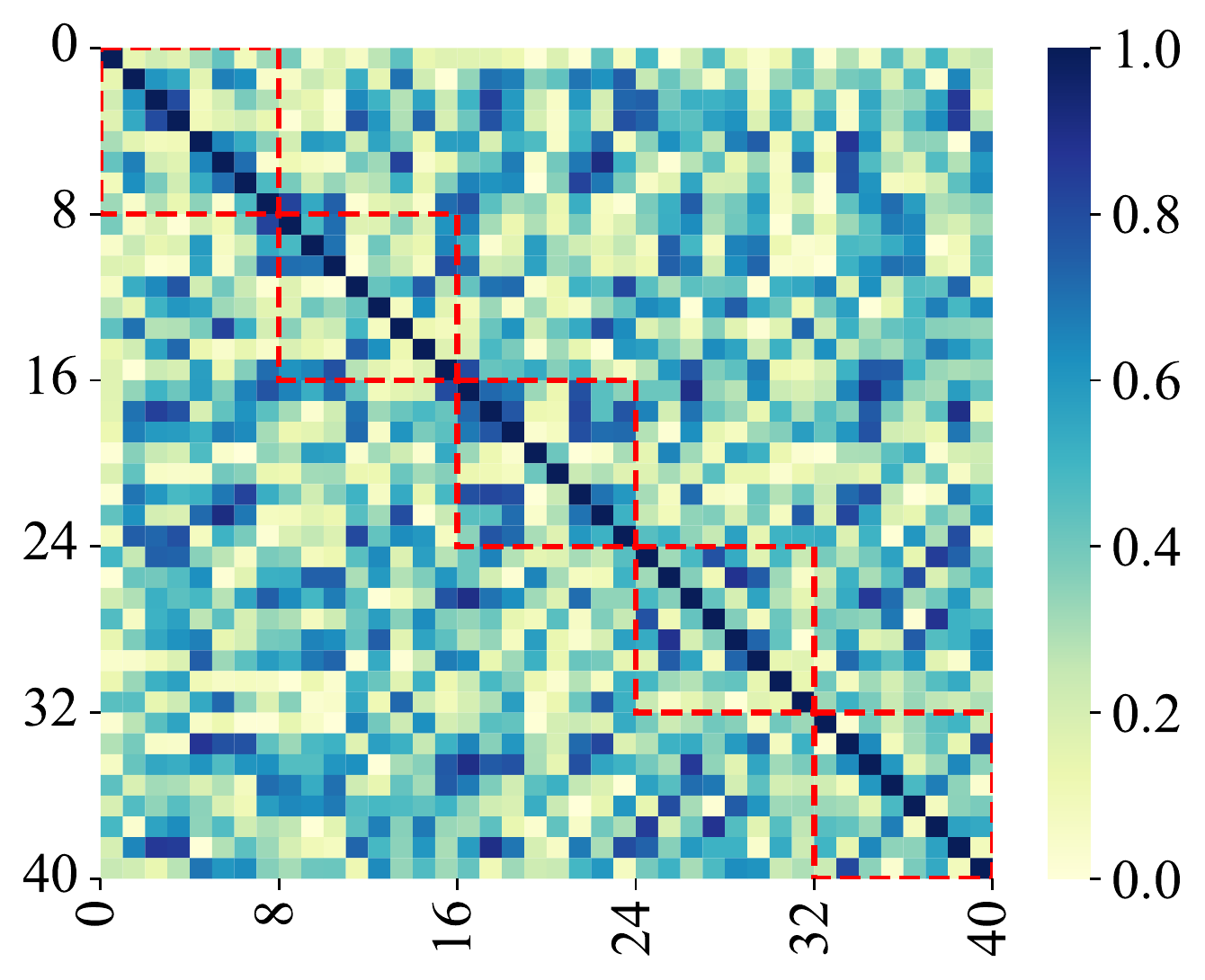}}\label{fig:feat_vgae}			
	\hfil
	\subfloat[w/o MI]{\includegraphics[width=0.24\linewidth]{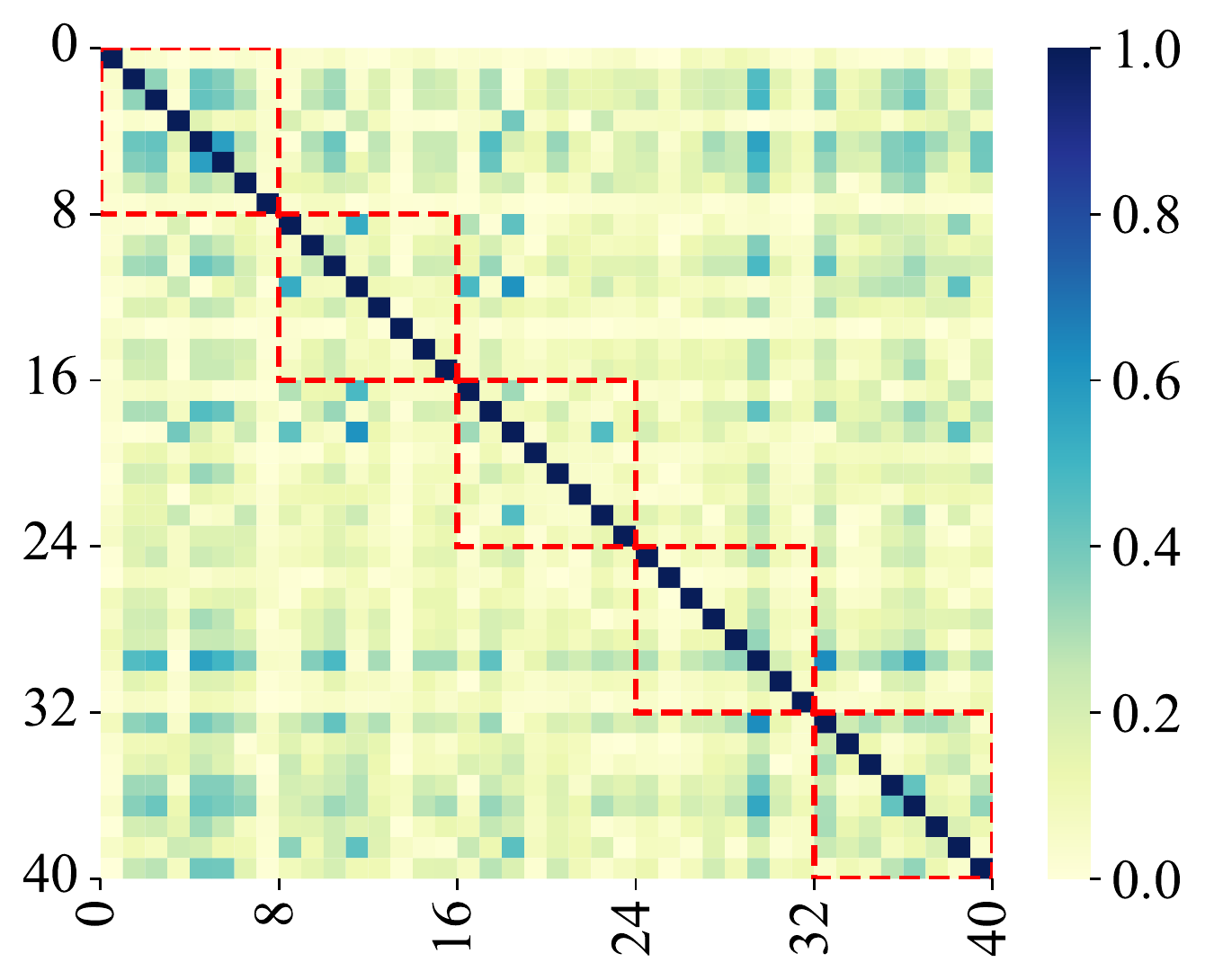}}\label{fig:feat_disengcn}
	\hfil	
	\subfloat[VDGAE]{\includegraphics[width=0.24\linewidth]{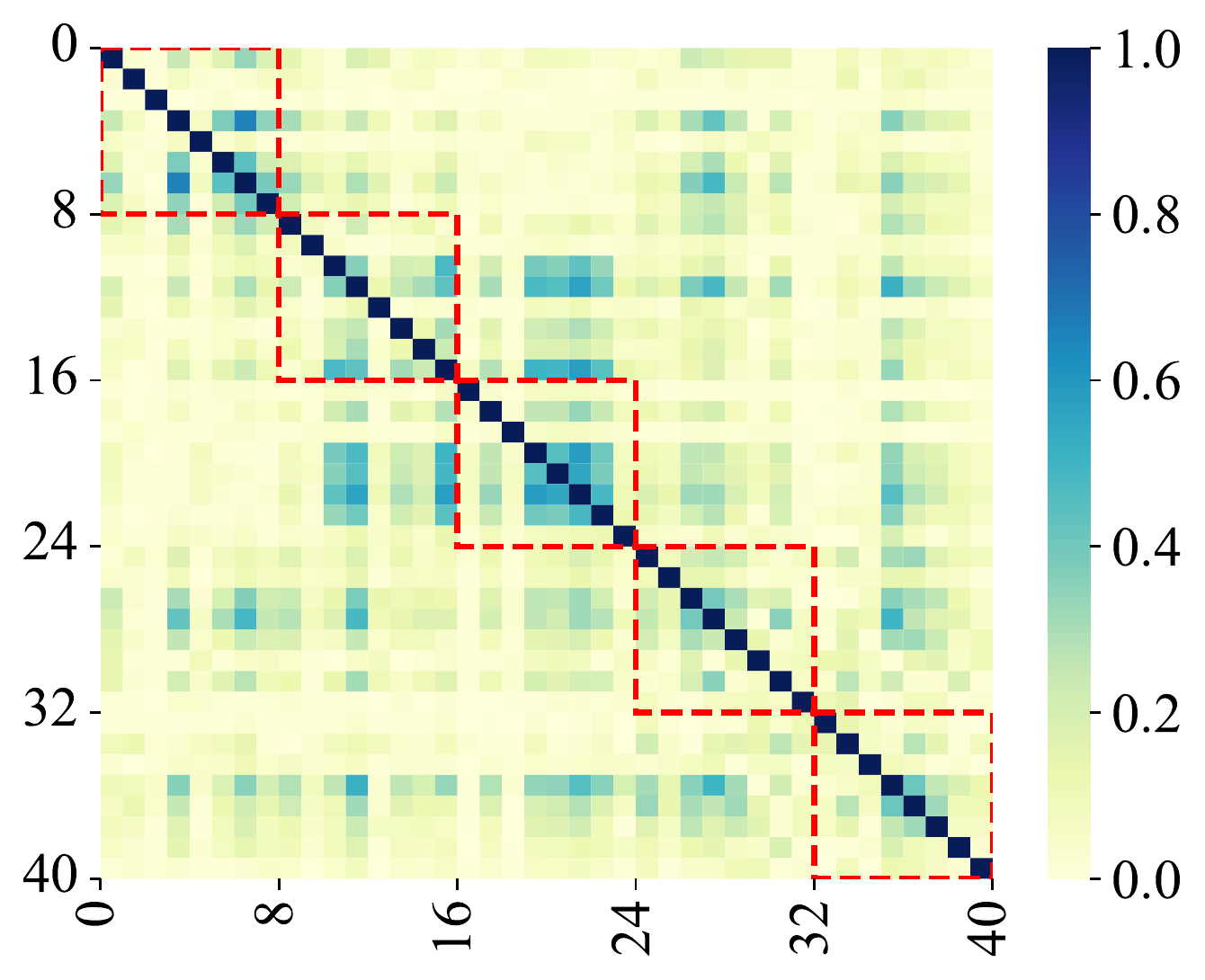}}\label{fig:feat_vdgae}		
	\caption{The correlation of the node embedding representations on the synthetic dataset. Compared with other models, VDGAE exhibits obvious diagonal blocks; in the regions except for the diagonal block,  VDGAE shows weak correlations.}
	\label{fig:disen}
	\vspace{-1em}
\end{figure*}
\begin{figure}[t!]
	\centering
	\subfloat[Raw features]{\includegraphics[width=0.32\linewidth]{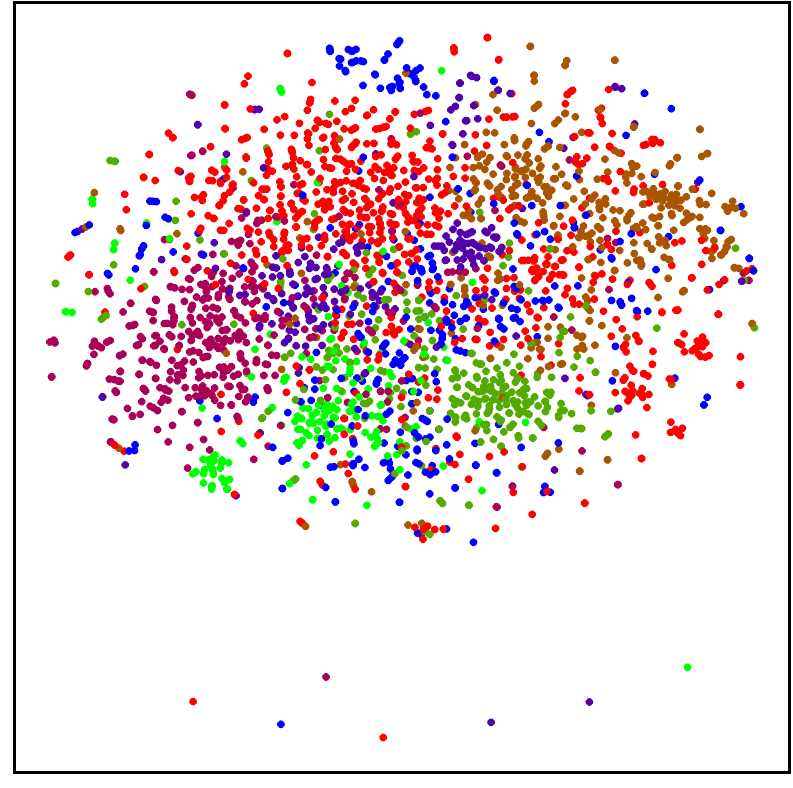}}\label{fig:raw}			
	\hfil
	\subfloat[VGAE]{\includegraphics[width=0.32\linewidth]{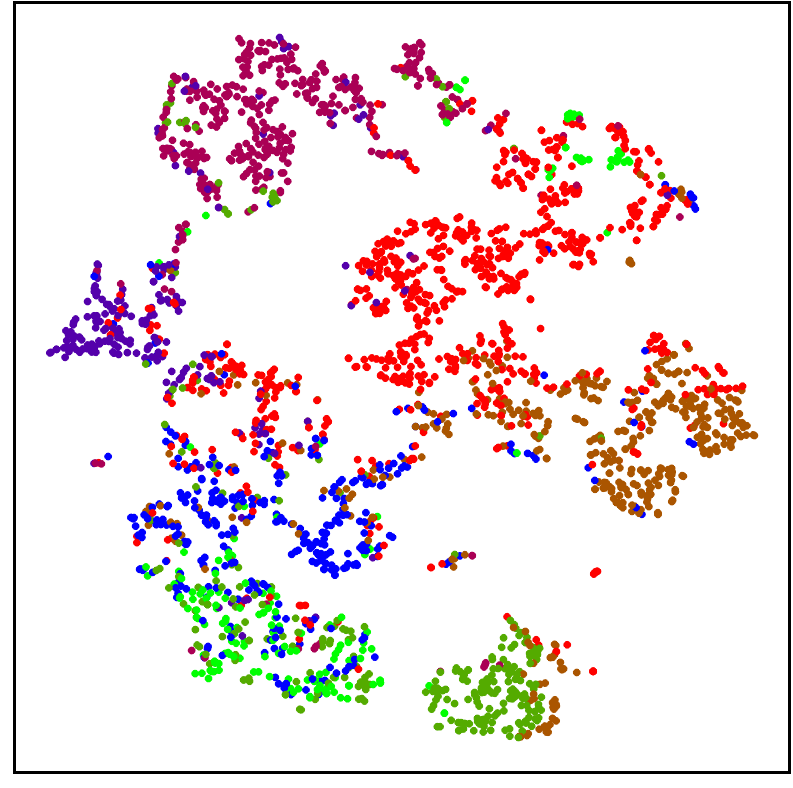}}\label{fig:vgae}			
	\hfil
	\subfloat[ARVGE]{\includegraphics[width=0.32\linewidth]{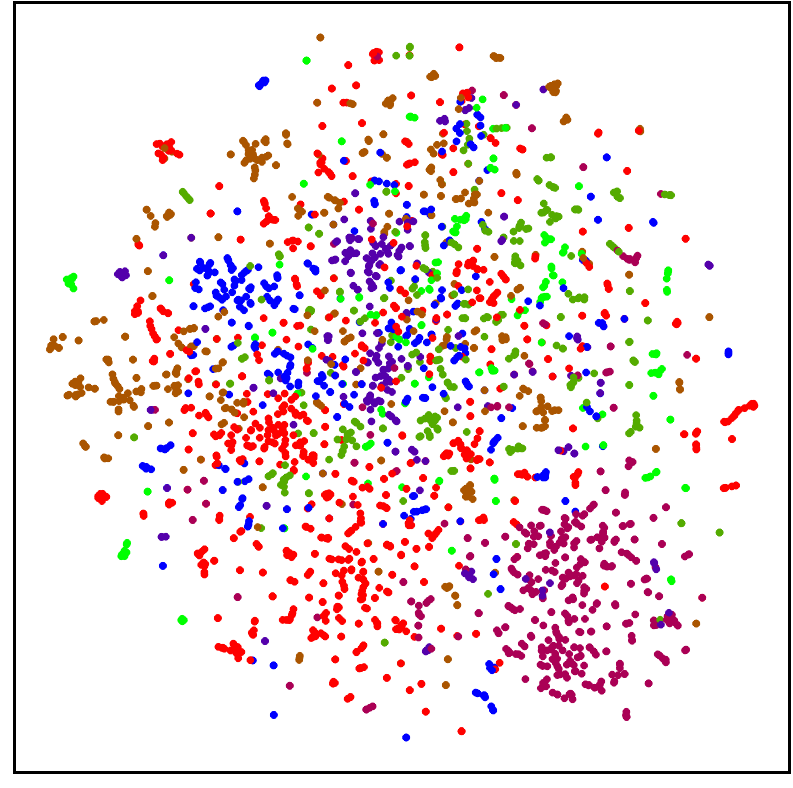}}\label{fig:argva}			
	\hfil
	
	\subfloat[VGNAE]{\includegraphics[width=0.32\linewidth]{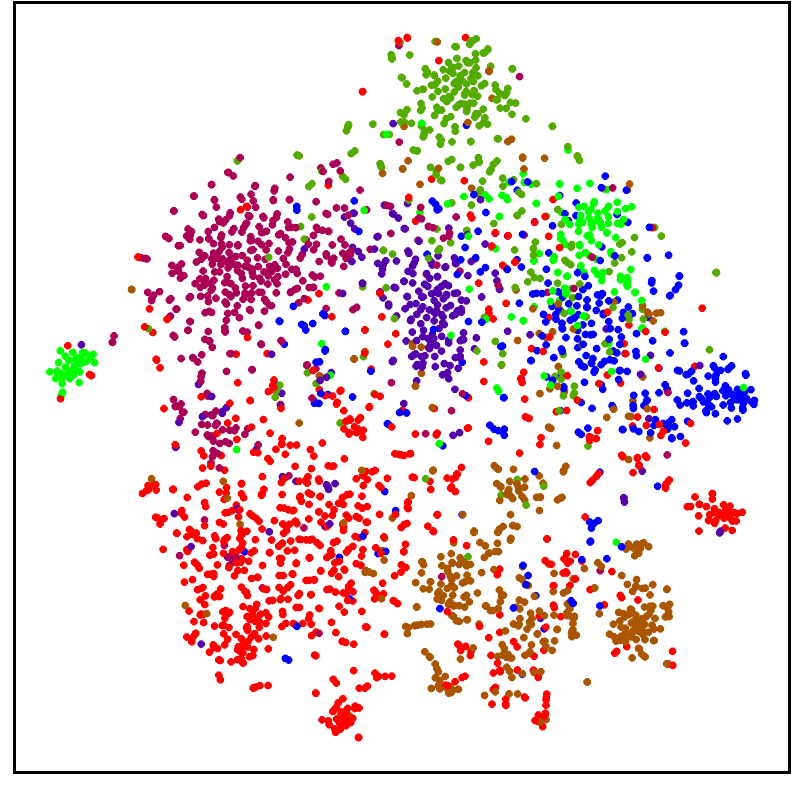}}\label{fig:vgnae}			
	\hfil	
	\subfloat[w/o MI]{\includegraphics[width=0.32\linewidth]{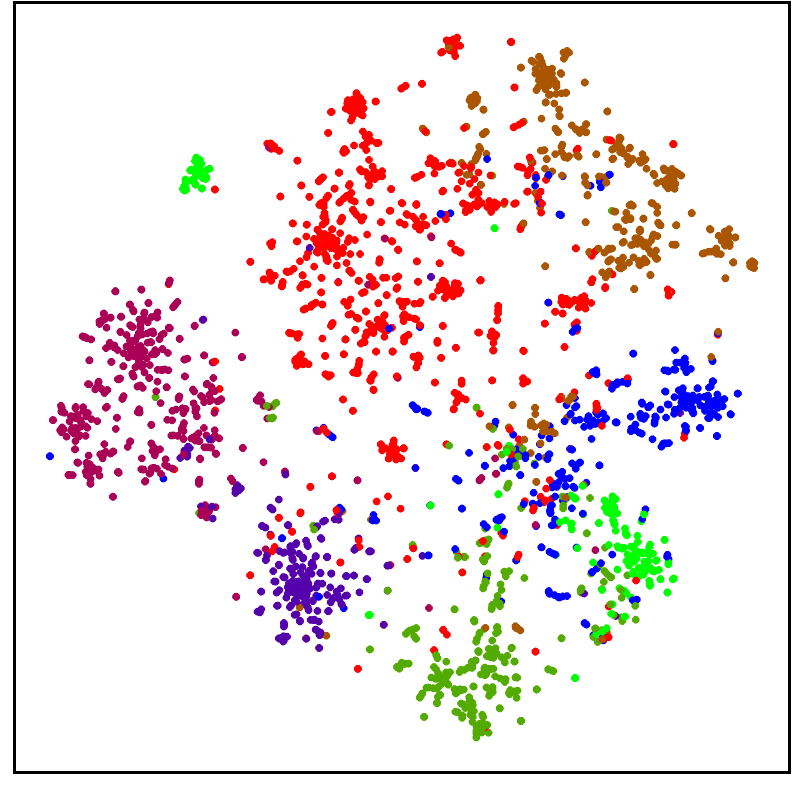}}\label{fig:disengcn}
	\hfil	
	\subfloat[VDGAE]{\includegraphics[width=0.32\linewidth]{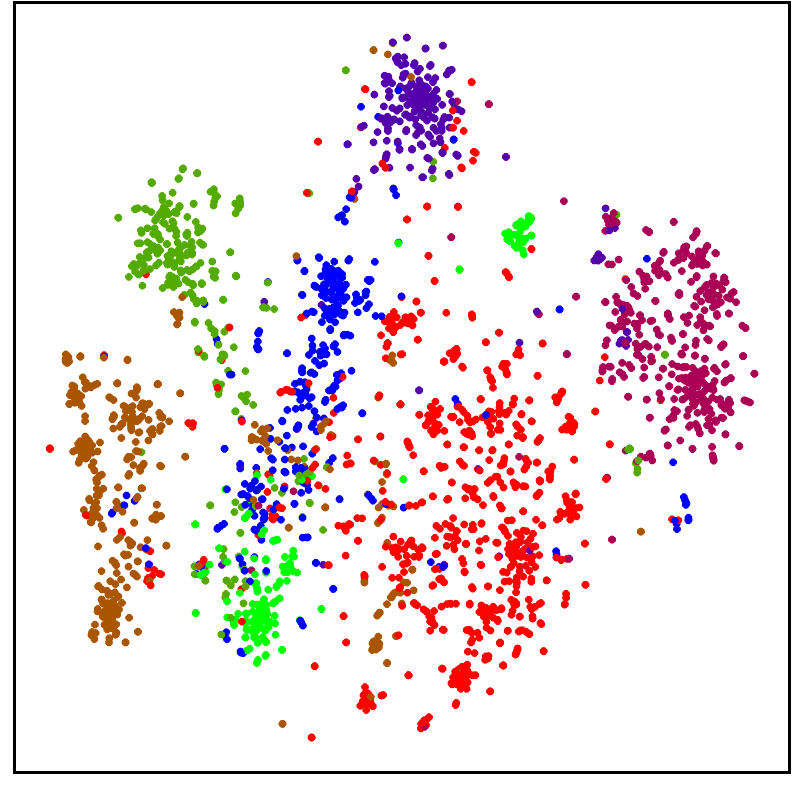}}\label{fig:vdgae}		
	\caption{Visualizations on the Cora dataset.}
	\label{fig:visual}
	\vspace{-1em}
\end{figure}
\subsection{Disentanglement Analysis}
Furthermore, to give a qualitative insight into the disentanglement mechanism, we generate a synthetic dataset with known latent factors and visualize the correlation of the node embedding representations learned by VDGAE on the synthetic dataset.

\subsubsection{Experimental Setup}
The synthetic dataset is generated using the stochastic block model \cite{holland1983stochastic} implemented in the networkx \cite{SciPyProceedings_11} library.  We assume that the dataset consists of $K$ communities, where each community corresponds to a latent factor. The latent factor is defined as the probability $p$ of establishing links between two nodes within the same community. To ensure graph connectivity, nodes from different communities are connected with the probability $q$.
Specifically, the synthetic dataset is created with 5 communities, with each community containing 500 nodes. To simulate the sparsity of real-world graphs, we select the non-repeatable probability $p$ for each community from  $\left \{  0.01, 0.02, \cdots, 0.09 \right \}$. The probability $q$ is carefully adjusted to maintain an average degree of the synthetic graph ranging from 18 to 20. We use the rows of the adjacency matrix as node features. The output dimension of each channel is $\Delta d=8$.

\subsubsection{Analysis}
Fig. \ref{fig:disen} summarizes the results of the correlation between the elements of the $5\times 8$-dimensional node embedding representations on the synthetic dataset. The node embedding representations in Fig. \ref{fig:disen}(a) is learned by a linear model, and the model structure is actually consistent with the structure of the second variant ($w/o\ disent$)  in the ablation study. Fig. \ref{fig:disen}(b), (c), (d) correspond to the correlation of the node embedding representations learned by VGAE, the first variant ($w/o\ MI$), and VDGAE respectively. In Fig. \ref{fig:disen}, we have the following important findings.
\begin{itemize}
	\item{Diagonal blocks: Compared with the other three methods, Fig. \ref{fig:disen}(d) exhibits five diagonal blocks, indicating that (1) VDGAE successfully distinguishes five different latent factors, and (2) the channels of VDGAE are likely to capture mutually exclusive information.}
	\item{Weak correlation: Except for the diagonal blocks, the other regions of Fig. \ref{fig:disen}(d) show weaker correlation, revealing that the proposed MI regularization strengthens the independence among different channels.}	
\end{itemize}

These findings suggest that the channels of VDGAE exhibit the capacity to identify unique latent factors related to different semantic information. Consequently, our proposed methods successfully learn disentangled node embeddings and capture latent factors that cause links, enabling link prediction results with a certain degree of credibility and interpretability.

\subsection{Visualization}
To present an intuitive sight, we visualize the raw features and the node embeddings learned by VGAE, ARVGE, VGNAE, the first variant ($w/o\ MI$), and VDGAE on the Cora dataset. The node representations are projected into a two-dimensional space by using t-SNE \cite{van2008visualizing}. The resulting visualizations, shown in Fig. \ref{fig:visual}, illustrate that the disentangled node embeddings learned by VDGAE are particularly effective at capturing the underlying information and producing a more meaningful layout of the graph data. These results highlight the superior performance of VDGAE in learning informative and disentangled representations of graph data.

\section{Conclusion}
In this paper, we propose a novel framework for link prediction with two variants, DGAE and VDGAE. We are pioneers in leveraging disentanglement strategy to enhance the accuracy and efficacy of link prediction. Our proposed framework effectively identifies latent factors in the graph and generates disentangled node embeddings, thus overcoming the problem of excessive interaction with irrelevant information. Additionally, we incorporate MI as a constraint to promote independence among the disentangled components and capture exclusive semantic information, leading to significant improvements in link prediction performance. Moreover, our methods achieve superior performance on link prediction compared to various strong baselines across multiple real-world datasets. Quantitative analysis validates the effectiveness and stability of our methods. We generate an synthetic dataset for conducting qualitative analysis, offering insights into link formation in complex networks. Furthermore, a particularly interesting direction for future work is to incorporate causal learning in order to learn the latent factors with causal semantics, resulting in causal disentangled node representations.

\vspace{-33pt}
\begin{IEEEbiography}[{\includegraphics[width=1in,height=1.25in,clip,keepaspectratio]{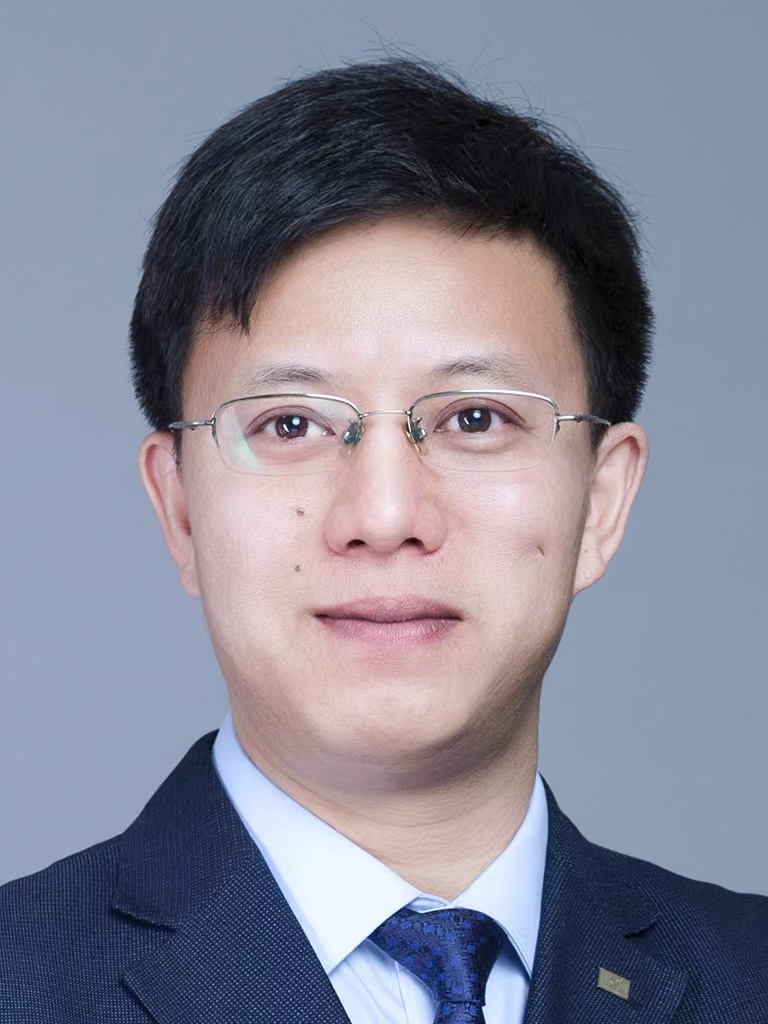}}]{Jun Fu}
	received a Ph.D. degree in mechanical engineering from Concordia University, Montreal, Quebec, Canada, in 2009. He was a Post-Doctoral  Researcher with the Department of Mechanical Engineering, Massachusetts Institute of Technology (MIT), Cambridge, MA, USA, from 2010 to 2014. He is a Full Professor with Northeastern University, Shenyang, China. His current research is on dynamic optimization, switched systems and their applications. 	
	
	Dr. Fu received the 2018 Young Scientist Award in Science issued by the Ministry of Education of China (the first awardee in Chinese Control Community). He is currently Associate Editors of Control Engineering Practice, the IEEE Transactions on Industrial Informatics, and the IEEE Transactions on Neural Networks and Learning Systems.
\end{IEEEbiography}

\begin{IEEEbiography}[{\includegraphics[width=1in,height=1.25in,clip,keepaspectratio]{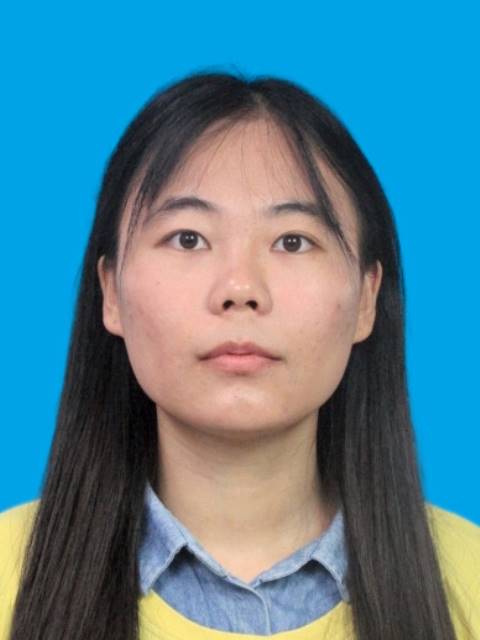}}]{Xiaojuan Zhang}
	received the B.S. degree in automation from Northeastern University, Shenyang, China, in 2021. She is currently pursuing the M.S. degree in control science and engineering with the State Key Laboratory of Synthetical Automation for Process Industries, Northeastern University, Shenyang, China.
	
	Her current research interests cover graph representation learning,  complex network and deep learning.
\end{IEEEbiography}
\vspace{-33pt}
\begin{IEEEbiography}[{\includegraphics[width=1in,height=1.25in,clip,keepaspectratio]{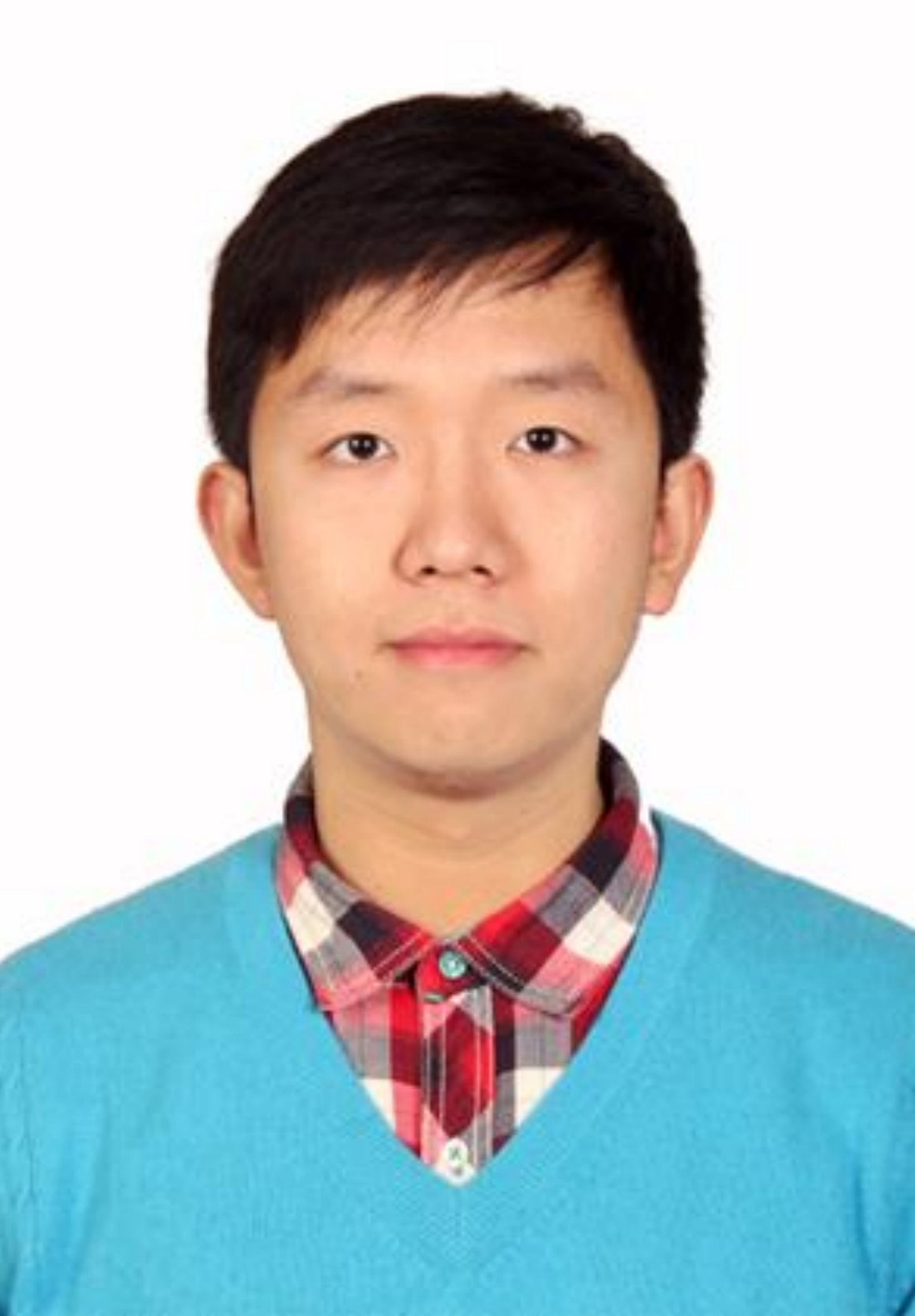}}]{Shuang Li}
	 received the Ph.D. degree in control science and engineering from the Department of Automation, Tsinghua University, Beijing, China, in 2018.
	
	He was a Visiting Research Scholar with the Department of Computer Science, Cornell University, Ithaca, NY, USA, from November 2015 to June 2016. He is currently an Associate Professor with the School of Computer Science and Technology, Beijing Institute of Technology, Beijing. His main research interests include machine learning and deep learning, especially in transfer learning and domain adaptation.
\end{IEEEbiography}
\vspace{-33pt}
\begin{IEEEbiography}[{\includegraphics[width=1in,height=1.25in,clip,keepaspectratio]{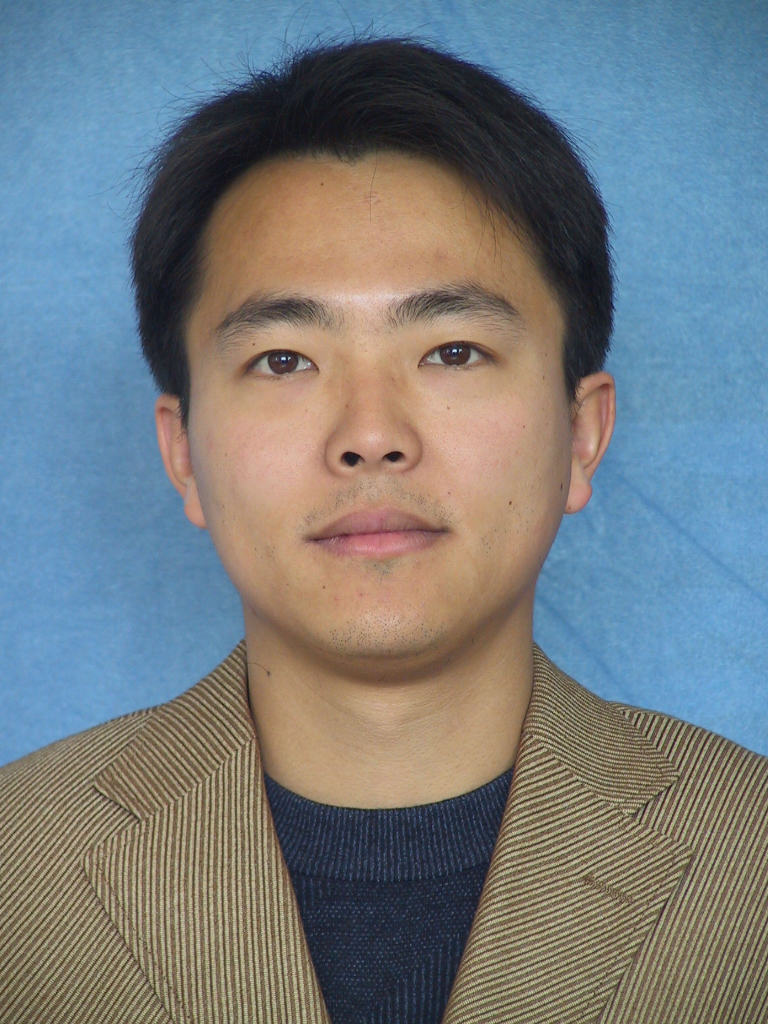}}]{Dali Chen}
	 received the Ph.D. degree in pattern recognition and intelligent system from Northeastern University, Shenyang, China, in 2008. He is a professor with the College of Information Science and Engineering, Northeastern University. His current research interests include computer vision and deep learning.
\end{IEEEbiography}
\balance

\begin{thebibliography}{1}
\bibitem{yang2013community}
J.~Yang, J.~McAuley, and J.~Leskovec, ``Community detection in networks with node attributes,'' 
in \emph{IEEE Intell. Conf. Data Mining}, 2013, pp. 1151--1156.

\bibitem{Zhang2010}
Y.~Zhang, Z.~Wang, and C.~Xia, ``Identifying key users for targeted marketing by mining online social network,'' 
in \emph{IEEE Intell. Conf. Adv. Inf. Networking Appl. Workshops}, 2010, pp. 644--649.

\bibitem{Lv2021}
G. Lv, Z. Hu, Y. Bi, and S. Zhang, ``Learning unknown from correlations: Graph neural network for inter-novel-protein interaction prediction,'' 
in \emph{Proc. Int. Joint Conf. Artif. Intell.}, 2021, pp. 3677--3683.

\bibitem{Karlebach2008}
G.~Karlebach and R.~Shamir, ``Modelling and analysis of gene regulatory networks,'' 
\emph{Nat. Rev. Mol. Cell Biol.}, vol.~9, no.~10, pp. 770--780, 2008.

\bibitem{Xia2022}
L. Xia, C. Huang, Y. Xu, P. Dai, and L. Bo, ``Multi-behavior graph neural networks for recommender system,'' 
\emph{IEEE Trans. Neural Networks Learn. Syst.}, pp. 1--15, 2022.

\bibitem{10.1145/3397271.3401137}
X.~Wang, H.~Jin, A.~Zhang, X.~He, T.~Xu, and T.-S. Chua, ``Disentangled graph collaborative filtering,'' 
in \emph{Proc. Int. ACM SIGIR Conf. Res. Develop. Inf. Retr.}, 2020, pp. 1001--1010.

\bibitem{Brin1998}
S.~Brin and L.~Page, ``The anatomy of a large-scale hypertextual web search engine,'' 
\emph{Comput. Networks ISDN Syst.}, vol.~30, no. 1-7, pp. 107--117, 1998.

\bibitem{Ji2022}
S.~Ji, S.~Pan, E.~Cambria, P.~Marttinen, and P.~S. Yu, ``A survey on knowledge graphs: Representation, acquisition, and applications,'' 
\emph{IEEE Trans. Neural Networks Learn. Syst.}, vol.~33, no.~2, pp. 494--514, 2022.

\bibitem{Lue2011}
L. L{\"u} and T. Zhou, ``Link prediction in complex networks: A survey,''
\emph{Physica A: statistical mechanics and its applications}, vol. 390, no. 6, pp. 1150--1170, 2011.

\bibitem{Barabasi1999}
A. L. Barab{\'a}si and R. Albert, ``Emergence of scaling in random networks,''
\emph{Science}, vol. 286, no. 5439, pp. 509--512, 1999.

\bibitem{Brin2012}
S. Brin and L. Page, ``Reprint of: The anatomy of a large-scale hypertextual web search engine,'' 
\emph{Computer networks}, vol. 56, no. 18, pp. 3825--3833, 2012.

\bibitem{Jeh2002}
G. Jeh and J. Widom, ``Simrank: a measure of structural-context similarity,''
in \emph{Proc. ACM SIGKDD Int. Conf. Knowl. Discovery Data Mining}, 2002, pp. 538--543.

\bibitem{Zhang2017}
M. Zhang and Y. Chen, ``Weisfeiler-lehman neural machine for link prediction,''
in \emph{Proc. ACM SIGKDD Int. Conf. Knowl. Discovery Data Mining}, 2017, pp. 575--583.

\bibitem{Adamic2003}
L. A. Adamic and E. Adar, ``Friends and neighbors on the web,'' 
\emph{Social networks}, vol. 25, no. 3, pp. 211--230, 2003.

\bibitem{Jaccard1902}
P. Jaccard, ``Distribution compar{\'e}e de la flore alpine dans quelques r{\'e}gions des alpes occidentales et orientales,'' 
\emph{Bulletin de la Murithienne}, no. 31, pp. 81--92, 1902.

\bibitem{Newman2001}
M. E. Newman, ``Clustering and preferential attachment in growing networks,''
\emph{Physical review E}, vol. 64, no. 2, p. 025102, 2001.

\bibitem{Zhou2009}
T. Zhou, L. L{\"u}, and Y. C. Zhang, ``Predicting missing links via local information,'' 
\emph{The European Physical Journal B}, vol. 71, pp. 623--630, 2009.

\bibitem{Katz1953}
L. Katz, ``A new status index derived from sociometric analysis,''
\emph{Psychometrika}, vol. 18, no. 1, pp. 39--43, 1953.

\bibitem{kingma2013auto}
D.~P. Kingma and M.~Welling, ``Auto-encoding variational bayes,'' 
in \emph{Proc. Int. Conf. Learn. Represent}, 2014.

\bibitem{Kipf2016}
T.~N. Kipf and M.~Welling, ``Variational graph auto-encoders,'' 
in \emph{Proc. NIPS Workshop}, 2016.

\bibitem{DBLP:conf/uai/DavidsonFCKT18}
T.~R. Davidson, L.~Falorsi, N.~D. Cao, T.~Kipf, and J.~M. Tomczak, ``Hyperspherical variational auto-encoders,'' 
in \emph{Proc. Conf. Uncert. Artif. Intell.}, 2018, pp. 856--865.

\bibitem{10.1145/2488388.2488393}
A. Ahmed, N. Shervashidze, S. Narayanamurthy, V. Josifovski, and A. J. Smola, ``Distributed large-scale natural graph factorization,'' 
in \emph{Proc. World Wide Web Conf.}, 2013, pp. 37--48.

\bibitem{10.1145/2806416.2806512}
S. Cao, W. Lu, and Q. Xu, ``{GraRep}: Learning graph representations with global structural information,'' 
in \emph{Proc. ACM Conf. Inf. Knowl. Manage}, 2015, pp. 891--900.

\bibitem{10.1145/2939672.2939751}
M. Ou, P. Cui, J. Pei, Z. Zhang, and W. Zhu, ``Asymmetric transitivity preserving graph embedding,'' 
in \emph{Proc. ACM SIGKDD Int. Conf. Knowl. Discovery Data Mining}, 2016, pp. 1105--1114.

\bibitem{10.1145/2623330.2623732}
B. Perozzi, R.~Al-Rfou, and S.~Skiena, ``{DeepWalk}: Online learning of social representations,'' 
in \emph{Proc. ACM SIGKDD Int. Conf. Knowl. Discovery Data Mining}, 2014, pp. 701--710.

\bibitem{Tang2015}
J.~Tang, M.~Qu, M.~Wang, M.~Zhang, J.~Yan, and Q.~Mei, ``{LINE:} large-scale information network embedding,'' 
in \emph{Proc. World Wide Web Conf.}, 2015, pp. 1067--1077.

\bibitem{10.1145/2939672.2939754}
A.~Grover and J.~Leskovec, ``Node2vec: Scalable feature learning for networks,'' 
in \emph{Proc. ACM SIGKDD Int. Conf. Knowl. Discovery Data Mining}, 2016, pp. 855--864.

\bibitem{Ribeiro2017}
L.~F. Ribeiro, P.~H. Saverese, and D.~R. Figueiredo, ``struc2vec: Learning node representations from structural identity,'' 
in \emph{Proc. ACM SIGKDD Int. Conf. Knowl. Discovery Data Mining}, 2017, pp. 385--394.

\bibitem{kipf2017semisupervised}
T.~N. Kipf and M.~Welling, ``Semi-supervised classification with graph convolutional networks,'' 
in \emph{Proc. Int. Conf. Learn. Represent}, 2017.

\bibitem{Hamilton2017}
W.~Hamilton, Z.~Ying, and J.~Leskovec, ``Inductive representation learning on large graphs,'' 
in \emph{Proc. Adv. Neural Inf. Process. Syst}, 2017.

\bibitem{DBLP:conf/iclr/VelickovicCCRLB18}
P.~Velickovic, G.~Cucurull, A.~Casanova, A.~Romero, P.~Li{\`{o}}, and Y.~Bengio, ``Graph attention networks,'' 
in \emph{Proc. Int. Conf. Learn. Represent}, 2018.

\bibitem{spectral}
D. K. Hammond, P. Vandergheynst, and R. Gribonval, ``Wavelets on graphs via spectral graph theory,'' 
in \emph{Appl. Comput. Harmon. Anal.}, vol.~30, no.~2, pp. 129--150, 2011.

\bibitem{DBLP:conf/ijcai/PanHLJYZ18}
S.~Pan, R.~Hu, G.~Long, J.~Jiang, L.~Yao, and C.~Zhang, ``Adversarially regularized graph autoencoder for graph embedding,'' 
in \emph{Proc. Int. Joint Conf. Artif. Intell.}, 2018, pp. 2609--2615.

\bibitem{Grover2019}
A.~Grover, A.~Zweig, and S.~Ermon, ``Graphite: Iterative generative modeling of graphs,'' 
in \emph{Proc. Int. Conf. Mach. Learn.}, 2019, pp. 2434--2444.

\bibitem{10.1145/3459637.3482215}
S.~J. Ahn and M.~Kim, ``Variational graph normalized autoencoders,'' 
in \emph{Proc. ACM Conf. Inf. Knowl. Manage}, 2021, pp. 2827--2831.

\bibitem{GonzalezGarcia2018}
A.~Gonzalez-Garcia, J.~van~de Weijer, and Y.~Bengio, ``Image-to-image translation for cross-domain disentanglement,'' 
in \emph{Proc. Adv. Neural Inf. Process. Syst}, 2018.

\bibitem{9241434}
Y.~Shen, C.~Yang, X.~Tang, and B.~Zhou, ``{InterFaceGAN}: Interpreting the disentangled face representation learned by gans,'' 
\emph{IEEE Trans. Pattern Anal. Mach. Intell.}, vol.~44, no.~4, pp. 2004--2018, 2022.

\bibitem{Ma2019}
J.~Ma, P.~Cui, K.~Kuang, X.~Wang, and W.~Zhu, ``Disentangled graph convolutional networks,'' 
in \emph{Proc. Int. Conf. Mach. Learn.}, 2019, pp. 4212--4221.

\bibitem{Liu2020}
Y.~Liu, X.~Wang, S.~Wu, and Z.~Xiao, ``Independence promoted graph disentangled networks,'' 
in \emph{Proc. AAAI Conf. Artif. Intell.}, 2020, pp. 4916--4923.

\bibitem{Zheng2021}
S.~Zheng, Z.~Zhu, Z.~Liu, S.~Ji, J.~Cheng, and Y.~Zhao, ``Adversarial graph disentanglement,'' 
\emph{arXiv preprint arXiv:2103.07295}, 2021.

\bibitem{Yang2020}
Y.~Yang, Z.~Feng, M.~Song, and X.~Wang, ``Factorizable graph convolutional networks,'' 
in \emph{Proc. Adv. Neural Inf. Process. Syst}, 2020, pp. 20286--20296.

\bibitem{Zhao2022}
T.~Zhao, X.~Zhang, and S.~Wang, ``Exploring edge disentanglement for node classification,'' 
in \emph{Proc. ACM Web Conf.}, 2022, pp. 1028--1036.

\bibitem{Li2021}
H.~Li, X.~Wang, Z.~Zhang, Z.~Yuan, H.~Li, and W.~Zhu, ``Disentangled contrastive learning on graphs,'' 
in \emph{Proc. Adv. Neural Inf. Process. Syst}, 2021, pp. 21872--21884.

\bibitem{10.1145/3459637.3482424}
J.~Wu, W.~Shi, X.~Cao, J.~Chen, W.~Lei, F.~Zhang, W.~Wu, and X.~He, ``{DisenKGAT}: Knowledge graph embedding with disentangled graph attention network,'' 
in \emph{Proc. ACM Conf. Inf. Knowl. Manage}, 2021, pp. 2140--2149.

\bibitem{Yu2022}
R.~Yu, S.~Gao, J.~Yu, M.~Zhao, T.~Xu, J.~Gao, H.~Liu, and X.~Li, ``Knowledge graph embedding with direct and disentangled neighborhood representation attention network,'' 
in \emph{Proc. Knowl. Sci. Eng. Manage.}, 2022, pp. 281--294.

\bibitem{10.1145/3534678.3539453}
Y.~Geng, J.~Chen, W.~Zhang, Y.~Xu, Z.~Chen, J.~Z.~Pan, Y.~Huang, F.~Xiong, and H.~Chen, ``Disentangled ontology embedding for zero-shot learning,'' 
in \emph{Proc. ACM SIGKDD Int. Conf. Knowl. Discovery Data Mining}, 2022, pp. 443--453.

\bibitem{10.1145/3340531.3411996}
Y.~Wang, S.~Tang, Y.~Lei, W.~Song, S.~Wang, and M.~Zhang, ``{DisenHAN}: Disentangled heterogeneous graph attention network for recommendation,'' 
in \emph{Proc. ACM Conf. Inf. Knowl. Manage}, 2020, pp. 1605--1614.

\bibitem{Liu2022}
Y.~Liu, M.~Jin, S.~Pan, C.~Zhou, Y.~Zheng, F.~Xia, and P.~Yu, ``Graph self-supervised learning: A survey,'' 
\emph{IEEE Trans. Knowl. Data Eng.}, 2022.

\bibitem{Cheng2020}
P.~Cheng, W.~Hao, S.~Dai, J.~Liu, Z.~Gan, and L.~Carin, ``{CLUB}: A contrastive log-ratio upper bound of mutual information,''
in \emph{Proc. Int. Conf. Mach. Learn.}, 2020, pp. 1779--1788.

\bibitem{USAir}
V.~Batagelj and A.~Mrvar, ``Pajek datasets,'' 2006. [Online]. Available:
\url{http://vlado.fmf.uni-lj.si/pub/networks/data/}

\bibitem{newman2006finding}
M.~E. Newman, ``Finding community structure in networks using the eigenvectors of matrices,'' 
\emph{Physical review E}, vol.~74, no.~3, p. 036104, 2006.

\bibitem{Ackland2005}
R.~Ackland \emph{et~al.}, ``Mapping the us political blogosphere: Are conservative bloggers more prominent?'' 
in \emph{BlogTalk Downunder 2005 Conf.}, Sydney, 2005.

\bibitem{VonMering2002}
C.~Von~Mering, R.~Krause, B.~Snel, M.~Cornell, S.~G. Oliver, S.~Fields, and P.~Bork, ``Comparative assessment of large-scale data sets of
protein--protein interactions,'' 
\emph{Nature}, vol. 417, no. 6887, pp. 399--403, 2002.

\bibitem{Watts1998}
D.~J. Watts and S.~H. Strogatz, ``Collective dynamics of ‘small-world’networks,'' 
\emph{Nature}, vol. 393, no. 6684, pp. 440--442, 1998.

\bibitem{Spring2002}
N.~Spring, R.~Mahajan, and D.~Wetherall, ``Measuring ISP topologies with rocketfuel,'' 
\emph{ACM SIGCOMM}, vol.~32, no.~4, pp. 133--145, 2002.

\bibitem{Zhang2018a}
M.~Zhang, Z.~Cui, S.~Jiang, and Y.~Chen, ``Beyond link prediction: Predicting hyperlinks in adjacency space,'' 
in \emph{Proc. AAAI Conf. Artif. Intell.}, 2018.

\bibitem{Yang2016}
Z.~Yang, W.~Cohen, and R.~Salakhudinov, ``Revisiting semi-supervised learning with graph embeddings,'' 
in \emph{Proc. Int. Conf. Mach. Learn.}, 2016, pp. 40--48.

\bibitem{Zhu2020}
J.~Zhu, Y.~Yan, L.~Zhao, M.~Heimann, L.~Akoglu, and D.~Koutra, ``Beyond homophily in graph neural networks: Current limitations and effective designs,'' 
in \emph{Proc. Adv. Neural Inf. Process. Syst}, 2020, pp. 7793--7804.

\bibitem{Pei2020Geom-GCN:}
H.~Pei, B.~Wei, K.~C.-C. Chang, Y.~Lei, and B.~Yang, ``Geom-GCN: Geometric graph convolutional networks,'' 
in \emph{Proc. Int. Conf. Learn. Represent}, 2020.

\bibitem{shchur2018pitfalls}
O.~Shchur, M.~Mumme, A.~Bojchevski, and S.~G{\"u}nnemann, ``Pitfalls of graph neural network evaluation,'' 
\emph{arXiv preprint arXiv:1811.05868}, 2018.

\bibitem{Fey/Lenssen/2019}
M.~Fey and J.~E. Lenssen, ``Fast graph representation learning with {PyTorch Geometric},'' 
in \emph{Proc. ICLR Workshop}, 2019.

\bibitem{Shervashidze2011}
N.~Shervashidze, P.~Schweitzer, E.~J. Van~Leeuwen, K.~Mehlhorn, and K.~M. Borgwardt, ``Weisfeiler-lehman graph kernels.'' 
\emph{Journal of Machine Learning Research}, vol.~12, no.~9, 2011.

\bibitem{tang2011leveraging}
L.~Tang and H.~Liu, ``Leveraging social media networks for classification,''
\emph{Data Min. Knowl. Discovery}, vol.~23, pp. 447--478, 2011.

\bibitem{DBLP:conf/pakdd/MavromatisK21}
C.~Mavromatis and G.~Karypis, ``{Graph InfoClust}: Maximizing coarse-grain mutual information in graphs,'' 
in \emph{Proc. Pacific–Asia Conf. Adv. Knowl. Discovery Data Mining}, 2021, pp. 541--553.

\bibitem{DBLP:conf/iclr/VelickovicFHLBH19}
P.~Velickovic, W.~Fedus, W.~L. Hamilton, P.~Li{\`{o}}, Y.~Bengio, and R.~D. Hjelm, ``Deep graph infomax,''
in \emph{Proc. Int. Conf. Learn. Represent}, 2019.

\bibitem{Zhang2018}
M.~Zhang and Y.~Chen, ``Link prediction based on graph neural networks,'' 
in \emph{Proc. Adv. Neural Inf. Process. Syst}, 2018, pp. 5171--5181.

\bibitem{DBLP:journals/corr/KingmaB14}
D.~P. Kingma and J.~Ba, ``Adam: {A} method for stochastic optimization,'' 
in \emph{Proc. Int. Conf. Learn. Represent}, 2015, pp. 7--9.

\bibitem{holland1983stochastic}
P.~W. Holland, K.~B. Laskey, and S.~Leinhardt, ``Stochastic blockmodels: First steps,'' 
\emph{Social Netw.}, vol.~5, no.~2, pp. 109--137, 1983.

\bibitem{SciPyProceedings_11}
A.~A. Hagberg, D.~A. Schult, and P.~J. Swart, ``Exploring network structure, dynamics, and function using networkx,'' 
in \emph{Proc. Python Sci. Conf.}, 2008, pp. 11--15.

\bibitem{van2008visualizing}
L.~Van~der Maaten and G.~Hinton, ``Visualizing data using t-sne,'' 
\emph{J. Mach. Learn. Res.}, vol.~9, no.~11, pp. 2579--2605, 2008.	
	
\end{thebibliography}
\end{document}